\documentclass{article}

\usepackage[preprint]{neurips_2023}
\setcitestyle{numbers,square}

\usepackage[utf8]{inputenc} 
\usepackage[T1]{fontenc}    
\usepackage{hyperref}       
\usepackage{url}            
\usepackage{graphicx}
\usepackage{multirow}
\usepackage{caption, subcaption}
\usepackage{booktabs}  
\usepackage[dvipsnames]{xcolor}
\usepackage{amssymb}  
\usepackage{amsmath}  
\usepackage{wrapfig}

\usepackage{algorithm}
\usepackage{algpseudocode}
\usepackage{float}
\usepackage{color}
\usepackage{colortbl}
\usepackage{rotating}
\usepackage{times}
\usepackage{epsfig}
\usepackage{enumitem}
\usepackage{makecell}
\usepackage{lineno}
\usepackage{tabularx}
\usepackage{xcolor}
\usepackage{multirow}
\usepackage{graphicx}
\usepackage{hhline}
\usepackage{textcomp,booktabs}
\usepackage{caption}
\usepackage{stmaryrd}
\usepackage[mathscr]{eucal}
\usepackage{lineno}
\usepackage[export]{adjustbox}
\usepackage{comment}
\usepackage{multicol,lipsum}
\usepackage{amsfonts}
\usepackage{arydshln}

\def\1{mathbb{1}}

\def\c{{\bf c}}

\def\e{{\bf e}}

\def\f{{\bf f}}

\def\p{{\bf p}}

\def\x{{\bf x}}

\def\z{{\bf z}}

\def\w{{\bf w}}
\def\0{{\bf 0}}
\def\1{{\bf 1}}

\definecolor{purple}{rgb}{0.56,0.27,0.68}
\definecolor{red}{rgb}{0.95,0.4,0.4}
\definecolor{purered}{rgb}{1,0,0}
\definecolor{blue}{rgb}{0.4,0.4,0.95}
\definecolor{darkblue}{rgb}{0,0,0.8}
\definecolor{grey}{rgb}{0.6,0.6,0.6}
\definecolor{col1}{RGB}{232, 161, 148}
\definecolor{col2}{RGB}{148, 187, 232}
\definecolor{col3}{RGB}{206, 239, 255}
\definecolor{lightgrey}{rgb}{0.85,0.85,0.85}
\definecolor{lightlightgrey}{rgb}{0.9,0.9,0.9}
\definecolor{verylightBG}{rgb}{0.9,0.99,0.99}
\definecolor{darkgreen}{rgb}{0.3, 0.75, 0.3}

\title{Improving Knowledge Distillation via Regularizing \\ 
Feature Norm and Direction}

\author{%
  Yuzhu Wang$^{\dagger}$~~
  Lechao Cheng$^{\dagger}$\thanks{Corresponding author: \texttt{chenglc@zhejianglab.com}}~~~
  Manni Duan$^{\dagger}$~~
  Yongheng Wang$^{\dagger}$~~
  Zunlei Feng$^{\ddagger}$~~
  Shu Kong$^{\mathsection}$\\
  $^{\dagger}$ Zhejiang Lab \\
  $^{\ddagger}$ Zhejiang University \\
  $^{\mathsection}$ Texas A\&M University 
}

\begin{document}

\maketitle

\begin{abstract}

Knowledge distillation (KD) exploits a large well-trained model (i.e., {\tt teacher}) to train a small {\tt student} model on the same dataset for the same task.
Treating {\tt teacher} features as knowledge, prevailing methods of knowledge distillation train {\tt student} by aligning its features with the {\tt teacher}'s, e.g., by minimizing the KL-divergence between their logits or L2 distance between their intermediate features. 
While it is natural to believe that better alignment of {\tt student} features to the {\tt teacher} better distills {\tt teacher} knowledge, simply forcing this alignment does not directly contribute to the {\tt student}'s performance, e.g., classification accuracy.
For example, minimizing the L2 distance between the penultimate-layer features (which are used to compute logits) does not necessarily help learn {\tt student}-classifier.
In this work, we propose to align {\tt student} features with class-mean of {\tt teacher} features, where class-mean naturally serves as a strong classifier.
To this end, we explore baseline techniques such as adopting the cosine distance based loss to encourage the similarity between {\tt student} features and their corresponding class-means of the {\tt teacher}.
Moreover, we train the {\tt student} to produce large-norm features, inspired by other lines of work (e.g., model pruning and domain adaptation), which find the large-norm features to be more significant.
Finally, we propose a rather simple loss term (dubbed ND loss) to simultaneously (1) encourage {\tt student} to produce large-\emph{norm} features, and (2) align the \emph{direction} of {\tt student} features and {\tt teacher} class-means. Experiments on standard benchmarks demonstrate that our explored techniques help existing KD methods achieve better performance, i.e., higher classification accuracy on ImageNet and CIFAR100 datasets, and higher detection precision on COCO dataset.
Importantly, our proposed ND loss helps the most, leading to the state-of-the-art performance on these benchmarks. The source code is available at \url{https://github.com/WangYZ1608/Knowledge-Distillation-via-ND}.

\end{abstract}

\section{Introduction}

Knowledge distillation (KD) is a well-studied technique to reduce the inference computation (e.g., running time and memory use) of a well-trained model.
Specifically, it aims to train a smaller model (called {\tt student}) by exploiting a larger one (called {\tt teacher}) which has been trained on the same dataset for the same task~\cite{hinton2015distilling}.
KD strives to train a {\tt student} to achieve better performance.
Compared to other related techniques such as pruning~\cite{ye2018rethinking} and quantization~\cite{han2015deep} (both of which also aim to reduce computation of a trained model and maintain its performance),
KD has the flexibility of using different architectures of the {\tt student}, which is an advantage in specific applications.

\begin{figure}[t]
\centering
    \includegraphics[width=13.9157cm, height=5.3578cm]{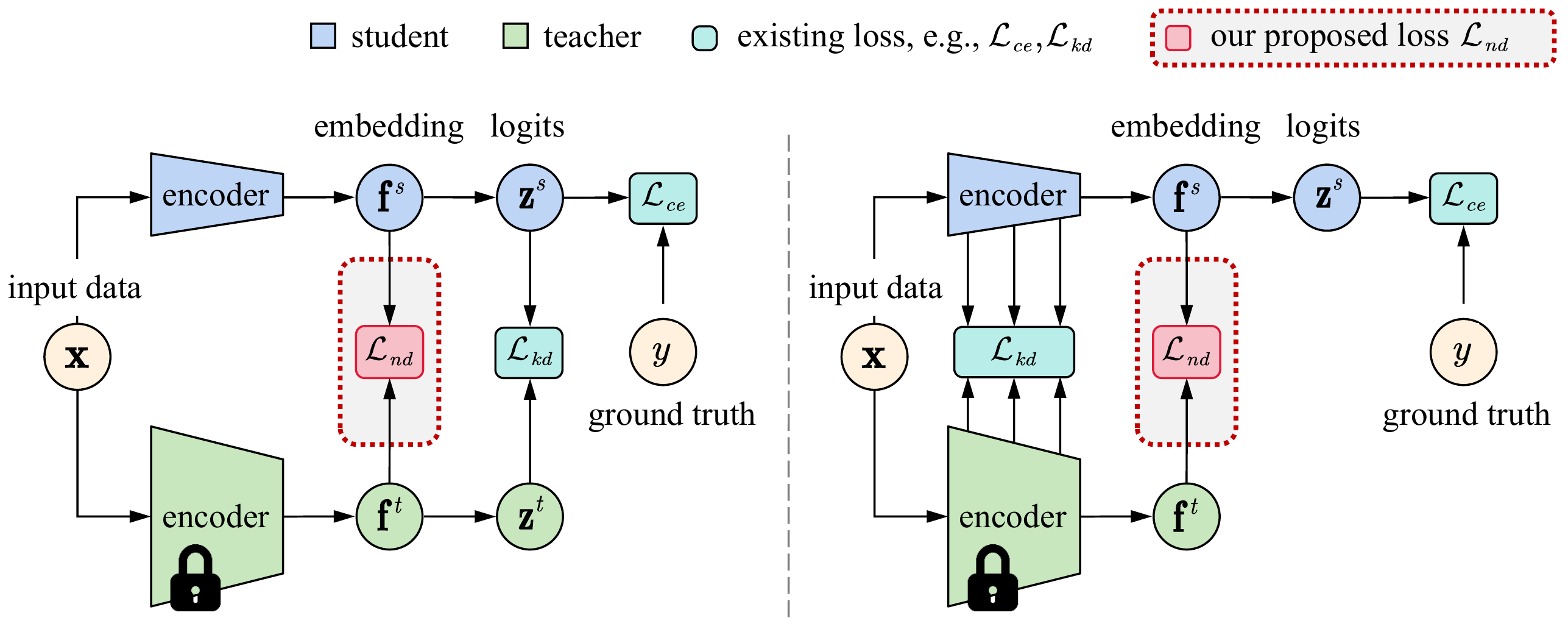}
    \\
    \vspace{-3mm}
    {\em logit distillation} \hspace{50mm} {\em feature distillation}
\caption{\small
Our main contribution is a simple loss, termed $\mathcal{L}_{nd}$, that regularizes the {\bf n}orm and {\bf d}irection of the {\tt student} features.
$\mathcal{L}_{nd}$ is applicable to different KD methods which are categorized into two types in the context of classification: (left) logit distillation that regularizes logits or softmax scores (e.g., KD~\cite{hinton2015distilling}
and DKD~\cite{zhao2022decoupled}),
and (right) feature distillation that regularizes features other than logits (e.g., ReviewKD~\cite{chen2021distilling}).
In this work, we particularly apply $\mathcal{L}_{nd}$ to the embedding feature, which is defined as the output at the penultimate layer before logits. 
Experiments show that learning with $\mathcal{L}_{nd}$ improves  existing KD methods, leading to the state-of-the-art benchmarking results for image classification (Table~\ref{tab:cifar100-sota}) and object detection (Table~\ref{tab:coco}).
}
\vspace{-8mm}
\label{fig:framwork}
\end{figure}

{\bf Status quo.}
Treating {\tt teacher} features as knowledge, KD methods train {\tt student} to distill this knowledge by encouraging its features to be similar to the {\tt teacher}'s.
In the context of classification, prevailing KD methods can be categorized into two types: logit distillation (Fig~\ref{fig:framwork}-left), and feature distillation (Fig~\ref{fig:framwork}-right).
Logit distillation trains the {\tt student} by minimizing the KL divergence between its logits and the {\tt teacher}’s for training data~\cite{hinton2015distilling, zhao2022decoupled}. It assumes that, if {\tt student} can produce logits more similar to {\tt teacher}’s, it should achieve better performance and approach {\tt teacher} performance . 
However, logit distillation methods do not make use of the full {\tt teacher} model, e.g., {\tt teacher} features at other layers are not exploited.
To exploit such, feature distillation methods train {\tt student} by encouraging its intermediate-layer features to be similar to the {\tt teacher}'s, e.g., minimizing the L2 distance between features~\cite{chen2021distilling, zagoruyko2016paying}.

{\bf Motivation.}
Despite the promising results of logit distillation and feature distillation methods, we point out that forcing the {\tt student} to produce similar logits or features to the {\tt teacher} does not directly serve the final task, e.g., classification.
For example, minimizing the L2 distance between the penultimate-layer features (which are used to compute logits) does not necessarily help learn better {\tt student}-classifier. 
Rather, {\tt student} features are presumably better guided by the {\tt teacher}-classifier.
Therefore, we propose to regularize {\tt student} features using class-mean of off-the-shelf features extracted by the {\tt teacher},
where class-mean naturally serves as a strong classifier~\cite{donahue2014decaf, sharif2014cnn, kong2021opengan}.
Moreover, 
other related lines of work (e.g., model pruning~\cite{ye2018rethinking} and domain adaptation~\cite{xu2019larger}) show the importance of large-norm features; a naively trained small-capacity model produces small-norm features (Fig.~\ref{fig:visual}).
These motivate us to train the {\tt student} to produce large-norm features.


{\bf Contributions.}
We make three main contributions in this work. 
First, we take a novel perspective to improve KD by regularizing {\tt student} to produce features that 
(1) have large \emph{norm}s and (2) are aligned with class-means constructed on off-the-shelf features of the {\tt teacher}.
Second, we extensively study baseline methods to achieve such regularizations.
We show that adopting them improves existing methods to achieve better KD performance, e.g., higher classification accuracy and higher detection precision of the {\tt student}.
Third, we propose a novel and simple loss that simultaneously regularizes feature \textbf{N}orm and \textbf{D}irection, termed as {\em ND loss}.
Experiments demonstrate that existing KD methods that additionally use the ND loss achieve better {\tt student} performance than using the baseline regularizers. 
For example, on the standard benchmark ImageNet~\cite{deng2009imagenet},
applying ND loss to KD~\cite{hinton2015distilling} achieves 72.53$\%$ classification accuracy (Table~\ref{tab:imagenet-largetea}), 
better than the original KD method (71.35$\%$),
using {\tt student} and {\tt teacher} architectures as ResNet-18 and ResNet-50, respectively, outperforming recently published methods (ReviewKD~\cite{chen2021distilling}: 71.10\%, DKD~\cite{zhao2022decoupled}: 71.87\%).

\section{Related Work\vspace{-1mm}}


\begin{figure}[t]
\centering
    \subcaptionbox{{\tt teacher} (Res56)}
    {\includegraphics[width = 0.3\textwidth, height=3.8cm]
    {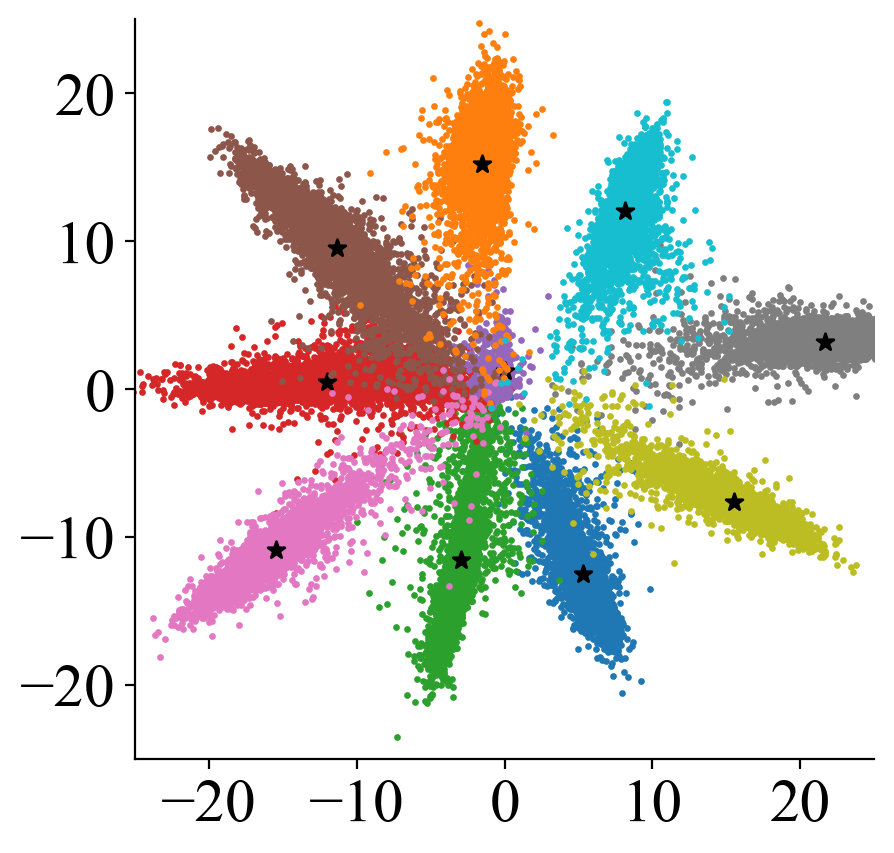}}
    \hfill
    \subcaptionbox{naive model (Res8)}
    {\includegraphics[width = 0.3\textwidth, height=3.8cm]
    {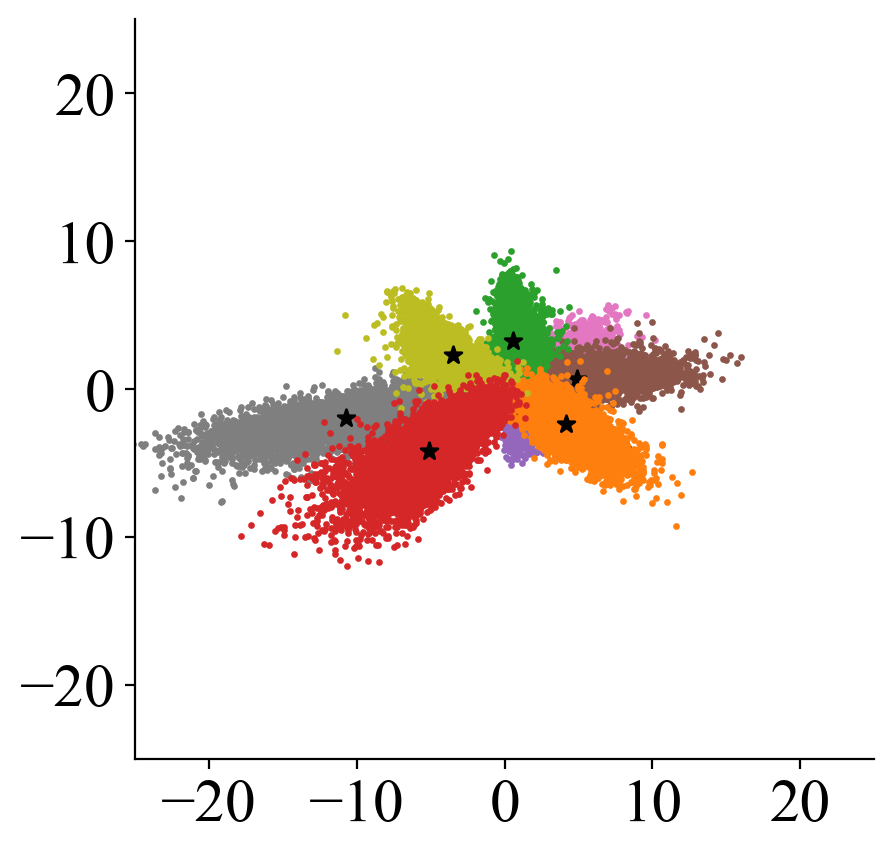}}
    \hfill
    \subcaptionbox{acc-vs-norm of Res8 {\tt student}}
    {\includegraphics[width = 0.33\textwidth, height=3.8cm]{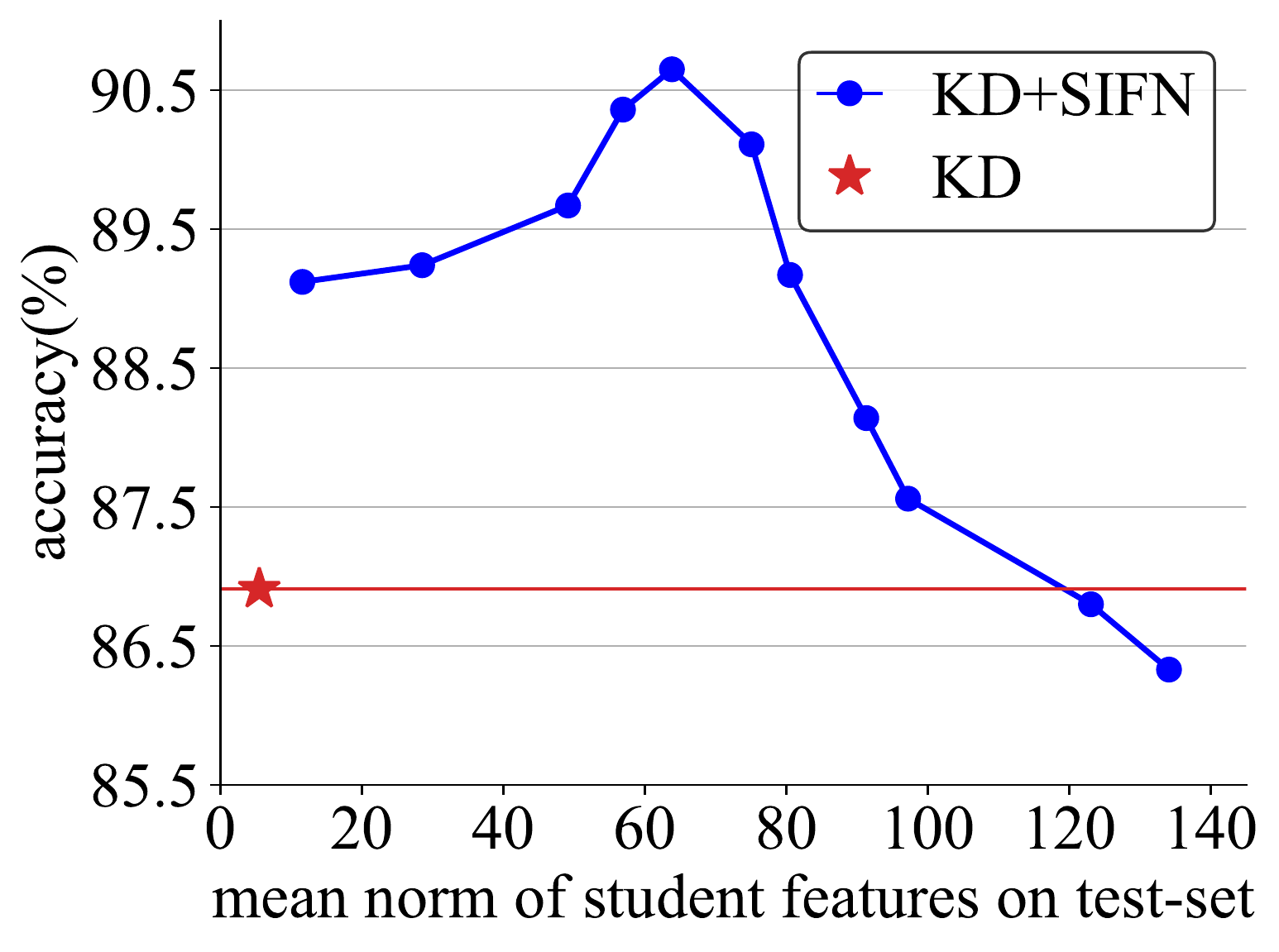}}
\vspace{-1mm}
\caption{\small
\textbf{Visualization of embedding features, and accuracy vs. feature norm.} 
We train {\tt teacher} (Res56 architecture), small-capacity Res8 model, and Res8 {\tt student} models on the CIFAR10 dataset.
For all the models, we purposely set the penultimate layer to produce 2D features for visualization in (a) and (b), 
where we visualize the training data as 2D points with colors indicating their class labels.
We mark the class center using $\star$.
Notably, the larger-capacity {\tt teacher} produce larger-norm features (a), while the small-capacity model produces small-norm features (b).
(c) Based on KD~\cite{hinton2015distilling}, training {\tt student} to produce larger-norm features using the method SIFN (Sec.~\ref{sec:FNR}) improves performance, quantitatively demonstrating the help of encouraging the {\tt student} to produce larger-norm features during training.
}
\vspace{-3mm}
\label{fig:visual}
\end{figure}

\textbf{Knowledge distillation} (KD) aims to train a small {\tt student} model by distilling knowledge of a well-trained large {\tt teacher} model.
The knowledge is delivered by features produced by the {\tt teacher} for training data.
Therefore, the key to KD is to align {\tt student} features to the {\tt teacher}'s.
The seminal KD method~\cite{hinton2015distilling} propose to train {\tt student} by aligning its logits with the {\tt teacher}'s, i.e., minimizing the Kullback-Leibler divergence (KL) between logits.
Other works improve KD by decoupling the KL loss into separate meaningful parts~\cite{zhao2022decoupled} or consider logits rankings~\cite{huang2022knowledge}.
As distilling logit knowledge alone may not be sufficient,  feature distillation aims to align more features at other layers~\cite{romero2014fitnets, zagoruyko2016paying, yim2017gift, heo2019comprehensive, passalis2018learning, park2019relational, tian2019contrastive, chen2021distilling, beyer2022knowledge}.
In this work, we take a different perspective to improve KD by encouraging the {\tt student} to produce features (at the penultimate layer before logits) that have large norm and are aligned with the direction of {\tt teacher} classifier.

{\bf Constructing classifiers using off-the-shelf features.}
Off-the-shelf features extracted from a well-trained model can be used to construct strong classifiers~\cite{donahue2014decaf, sharif2014cnn, kong2021opengan}.
One simple classifier is to compute class-mean  of training examples in the feature space, and uses such as the classifier~\cite{donahue2014decaf, sharif2014cnn, kong2021opengan}.
On the other hand, recent literature of pretrained large models~\cite{radford2021learning} shows that using off-the-shelf features and cosine similarity is a powerful classifier for zero-shot recognition. 
In this work, we propose to regularize {\tt student} features using class-mean of {\tt teacher} features.
We hypothesize that doing so helps learn better {\tt student} in terms of higher classification accuracy. 
Indeed, our experiments justify this hypothesis (Table~\ref{tab:ablation}).

\textbf{Large-norm feature matters.}
The literature of model pruning reports that features with smaller norms play a less informative role during the inference~\cite{ye2018rethinking}, so pruning elements or channels that produce small-norm features causes minimal performance drop.
The literature of domain adaptation~\cite{xu2019larger} reveals that the erratic discrimination of the target domain mainly stems from its much smaller feature norms w.r.t that of the source domain, and adopting a larger-norm constraint facilitates the more informative and transferable computation on the target domain. 
In our work, we empirically find that a small-capacity model produces features that tend to collide in the small-norm region (Fig.~\ref{fig:visual}b).
Therefore, we are motivated to train {\tt student} to produce large-norm features. 
Our experiments convincingly show that doing so leads to better KD performance (Fig.~\ref{fig:visual}c).


\section{Regularizing Feature Norm and Direction in Knowledge Distillation}

We describe notations with the KD background 
and motivate our study of regularizing feature norm and direction to improve KD.
Then, we introduce baselines, followed by the proposed ND loss.

\subsection{Notations and Background}

{\bf Notations.}
Without losing generality, we think of a classification neural network as two modules: a feature extractor $f(\cdot; \Theta)$, and a classifier $g(\cdot; \w)$, which are parameterized by $\Theta$ and $\w$, respectively.
For the {\tt teacher}, given input data $\x$,
we denote its embedding feature as $\f^t = f^t(\x; \Theta^t)$,
and the logits as $\z^t = g^t(\f^t; \w^t)$.
Similarly, the {\tt student} outputs the embedding features for $\x$ as $\f^s=f^s(\x; \Theta^s)$ and logits as $\z^s=g^s(\f^s; \w^s)$.
We compute softmax scores in a vector $\mathbf{q}^t= \text{softmax}(\mathbf{z}^t; \tau)$, where $\tau$ is a temperature (default 1 when training models from scratch).

Given the training set of $N$ examples belong to $C$ classes, $\x_i$ and its label $y_i$ (where $i=1,\dots,N$), we train a classification model (e.g., the {\tt teacher}) by minimizing the cross-entropy (CE) loss $\mathcal{L}_{ce}$ on all the training data.

{\bf Logit distillation}
trains the {\tt student} by transferring the {\tt teacher} knowledge using both the CE loss $\mathcal{L}_{ce}$ and a KD loss $\mathcal{L}_{kd}$.
The seminal work of KD~\cite{hinton2015distilling} uses KL divergence as the KD loss $\mathcal{L}_{kd}$,
i.e., $\mathcal{L}_{kd}= \frac{1}{N} \sum_{i=1}^N \mathbf{KL}(\mathbf{q}^t_i, \mathbf{q}^s_i)$. 


{\bf Feature distillation} distills {\tt teacher} knowledge by minimizing the difference of intermediate features at more layers other than the logits~\cite{zagoruyko2016paying, yim2017gift, chen2021distilling}.
A typical loss term is the L2 distance $\mathcal{L}_2$ between {\tt student} and {\tt teacher} features.\footnote{Features of {\tt student} and {\tt teacher} might have different dimensions.
This can be addressed by learning extra modules along with {\tt student} to project its features to the same dimension as {\tt teacher}'s~\cite{chen2021distilling}.}
For example, over the embedding features at the penultimate layer (before logits), we apply the loss $\mathcal{L}_{kd} = \frac{1}{N}\sum_{i=1}^N  \mathcal{L}_2(\mathbf{f}^s_i, \mathbf{f}^t_i)$ in addition to the CE loss $\mathcal{L}_{ce}$.
The final loss for KD can be written as
\begin{equation}\label{eq1}
    \mathcal{L} = \mathcal{L}_{ce}+\alpha \mathcal{L}_{kd}
\end{equation}
where $\alpha$ controls the significance of the KD loss $\mathcal{L}_{kd}$ depending on distillation choice: either logit distillation or feature distillation.




\subsection{Baseline Methods of Feature Norm and Direction Regularization}
\label{suboptimal}

Recall that we are motivated to regularize {\tt student} features during training: encouraging them to be large in norm and aligned with the class-mean of {\tt teacher} features. 
In this work, we focus on the embedding features $\f^s$ at the penultimate layer, which are directly used for classification.
We compute the class-mean of the $k^{th}$ class as $\c_k=\frac{1}{\vert {\cal I}_k \vert }\sum_{j \in {\cal I}_k} \mathbf{f}_j^t$,
where ${\cal I}_k$ is the set of indices of training examples belonging to class-$k$.
We now introduce simple techniques to regularize {\tt student} features using $\c_k$ in terms of feature norm and direction.

\subsubsection{Feature Norm Regularization}\label{sec:FNR}

\textbf{$\mathcal{L}_2$ distance}.
As shown by Fig.~\ref{fig:visual}b, small-capacity models such as  {\tt student} produce small-norm features.
To train the {\tt student} to produce larger-norm features, perhaps a naive method is to minimize the L2 distance between features of {\tt student} and {\tt teacher}:
\begin{equation}\label{eq:MSE}
    \mathcal{L}_{n} = \frac{1}{C}\sum_{k=1}^C \frac{1}{\vert{\cal I}_k\vert} \sum_{i\in {\cal I}_k}
    \Vert 
    \mathbf{f}^s_i - \mathbf{f}^t_i
    \Vert^2_2
\end{equation}
While minimizing Eq.~\ref{eq:MSE} is a common practice in feature distillation, it implicitly trains {\tt student} to produce features with norms approaching the corresponding larger-norm {\tt teacher} features.


\textbf{Stepwise increasing feature norms (SIFN).}
We now describe a loss to explicitly increase the norm of the {\tt student} features.
Inspired by \cite{xu2019larger}, 
we gradually increase the feature norm by minimizing:
\begin{equation}\label{eq:SFN}
    \mathcal{L}_{n} = \frac{1}{N}\sum_{i=1}^{N}\mathcal{L}_{2}\left( f^s(\x_i; \Theta^s_{previous}) + r, 
    f^s(\x_i; \Theta^s_{current})
\right)
\end{equation}
where $\Theta^s_{previous}$ and $\Theta^s_{current}$ are parameters of an early checkpoint and the current model being optimized, respectively;
$ r$ is a step size to increase the norm of {\tt student} features during training.

\subsubsection{Feature Direction Regularization}\label{sec:FDR}

\textbf{Cosine similarity.} 
We use a simple cosine similarity based loss term to regularize the feature direction of $\mathbf{f}^s_i$ according to its corresponding class-mean $\mathbf{c}_k$:
\begin{equation}\label{eq:cosine}
    \mathcal{L}_{d} =
    \frac{1}{C}\sum_{k=1}^C
    \frac{1}{\vert{\cal I}_k \vert} \sum_{i\in {\cal I}_k} (1 - \cos(\mathbf{f}_i^s, \mathbf{c}_k))
\end{equation}

\textbf{InfoNCE.}
Using the cosine similarity loss Eq.~\ref{eq:cosine} considers only paired examples and their corresponding class-mean.
Inspired by InfoNCE~\cite{oord2018representation}, we  also consider inter-class examples and class-means.
Therefore, we train {\tt student} by also minimizing:
\begin{equation}\label{eq:nce}
    \mathcal{L}_{d} = 
    \frac{1}{C}\sum_{k=1}^C
    \frac{1}{\vert{\cal I}_k \vert} \sum_{i\in {\cal I}_k}
    - \log 
    \frac{\exp \left( \cos (\mathbf{f}^s_i, \mathbf{c}_k \right))}
    {\sum\nolimits_{j=1}^C{\exp \left(\cos(\mathbf{f}^s_i, \mathbf{c}_j\right))}}
\end{equation}



\subsection{The Proposed ND Loss}

\begin{wrapfigure}{r}{0.3\textwidth}
\centering
\vspace{-7mm}
\includegraphics[width = 0.25\textwidth]{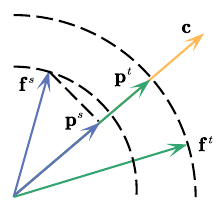}
\vspace{-2mm}
\caption{
\small
Illustration of notations used in our ND loss.}
\vspace{-2mm}
\label{fig:project} 
\end{wrapfigure}

For simplicity, we drop the subscript (i.e., the index of a  training example or class ID).
Let $\f^s$ and $\f^t$ be the embedding features of an input example $\x$ computed by {\tt student} and {\tt teacher}, respectively.
Based on $\x$'s ground-truth label $y$, we have its corresponding class-mean $\c$.
We compute the projection of $\f^s$ along the direction of $\c$: 
$\p^s = \f^s \cos(\f^s, \c)$.
We denote the unit vector $\e=\c/\Vert \c \Vert_2$,
and $\p^t = \e {\Vert\f^t\Vert_2 }$.
For physical meaning, please refer to  Fig.~\ref{fig:project}.

When the norm of $ \mathbf{f}^s$  is small, or its projection $\p^s$ has small norm, i.e., 
 $\Vert \p^s \Vert_2 < \Vert \f^t\Vert_2$, 
we encourage the {\tt student} to output larger-norm features and align them with the {\tt teacher} class-mean by minimizing
$\lVert \mathbf{p}^t - \mathbf{p}^s \rVert_2$.
Because the feature norms of different examples can vary by an order of magnitude (see Fig.~\ref{fig:visual}a), naively learning with the above can produce artificially large gradients from specific training data and negatively affect training.
Thus, we divide the above by $\Vert \f^t\Vert_2$, which is equivalent to $\Vert \p^t\Vert_2$:
\begin{equation}\label{eq:nd1}
\mathcal{L}_{nd} = 
    \frac{\lVert \mathbf{p}^t - \mathbf{p}^s \rVert_2}
    {\lVert \mathbf{f}^t\rVert_2}
= \frac{\lVert \p^t\rVert_2 - \Vert \p^s \rVert_2}
    {\lVert \mathbf{f}^t\rVert_2}
=1 - \frac{{\f^s} \cdot \mathbf{e}}
    {\lVert \mathbf{f}^t \rVert_2}
\end{equation}
Minimizing Eq.~\ref{eq:nd1} amounts to simultaneously (1) increasing the norm of $\f^s$ and (2) reducing the angular distance between $\f^s$ and the class-mean $\c$. 

When the norm of $ \mathbf{f}^s$, i.e.,  $\Vert \f^s \Vert_2 \ge \Vert \f^t\Vert_2$,
we do not need $\f^t$ to help increase feature norm for {\tt student}.
Instead, we use below to factor out the effect of feature norm :
\begin{equation}\label{eq:nd2}
\mathcal{L}_{nd} 
= 1 - \frac{{\f^s} \cdot \e}
    {\lVert \mathbf{f}^s \rVert_2}
\end{equation}
We merge Eq.~\ref{eq:nd1} and \ref{eq:nd2} and average over all training examples as our ND loss (dropping constant 1):
\begin{equation}\label{eq:nd3}
    \mathcal{L}_{nd} 
= - \frac{1}{C}\sum_{k=1}^C \frac{1}{\vert{\cal I}_k\vert} 
    \sum_{i \in {\cal I}_k} \frac{{\f^s_i} \cdot \e_k}
{\max \left\{ \lVert \mathbf{f}^s_i \rVert_2,  \lVert \mathbf{f}^t_i \rVert_2 \right\}}
\end{equation}

Compatible with existing KD methods, our ND loss $\mathcal{L}_{nd}$ can be used altogether with CE loss $\mathcal{L}_{ce}$ and KD loss $\mathcal{L}_{kd}$ to train {\tt student}: 
\begin{equation}
\label{eq:totalloss}
    \mathcal{L} = \mathcal{L}_{ce} + 
    \alpha\mathcal{L}_{kd} + 
    \beta\mathcal{L}_{nd} 
\end{equation}
$\alpha$ and $\beta$ are the weights for $\mathcal{L}_{kd}$ and $\mathcal{L}_{nd}$, respectively.
$\mathcal{L}_{kd}$ depends on the distillation method.
Otherwise stated, we study of $\mathcal{L}_{nd}$ with the seminal logit distillation method KD~\cite{hinton2015distilling}, so $\mathcal{L}_{kd}=\textbf{KL}$.

{\bf Remark.}
ND loss encourages {\tt student} to output larger-norm features during training. Importantly, {\tt student} feature norms can be larger than {\tt teacher}'s.
Moreover, ND loss directly minimizes the angular distance between {\tt student} features and the class-mean defined by the {\tt teacher}.
This is a desired property in terms of training the {\tt student} to achieve better classification accuracy.

\section{Experiments}

\begin{table}
    \centering
        \caption{\small Benchmarking results on the CIFAR100 dataset. Methods are reported with top-1 accuracy (\%). \textbf{++} means that we apply the proposed ND loss to existing approaches.
        Clearly, doing so improves performance over the original KD methods, and, importantly outperforms prior KD methods.
        }
        \renewcommand{\arraystretch}{1.2}
        \resizebox{\textwidth}{!}{%
        \begin{tabular}{c|ccc|ccc}
        \hline
        \multirow {3}{*}{Methods} & \multicolumn{3}{c}{Homogeneous architectures} & \multicolumn{3}{|c}{Heterogeneous architectures}\\ 
        \cline{2-7}
        ~ & ResNet-56 & WRN-40-2 & ResNet-32$\times$4 & ResNet-50 & ResNet-32$\times$4 & ResNet-32$\times$4 \\
        ~ & ResNet-20 & WRN-40-1 & ResNet-8$\times$4 & MobileNet-V2 & ShuffleNet-V1 & ShuffleNet-V2 \\
        \hline
        {\tt teacher} (T) & 72.34 & 75.61 & 79.42 & 79.34 & 79.42 & 79.42\\
        {\tt student} (S) & 69.06 & 71.98 & 72.50 & 64.60 & 70.50 & 71.82\\
        \hline
        \multicolumn{7}{c}{\textit{Feature distillation methods}}\\
        FitNet~\cite{romero2014fitnets}   & 69.21 & 72.24 & 73.50 & 63.16 & 73.59 & 73.54\\
        RKD~\cite{park2019relational}      & 69.61 & 72.22 & 71.90 & 64.43 & 72.28 & 73.21\\
        PKT~\cite{passalis2018probabilistic}	     & 70.34 & 73.45 & 73.64 & 66.52 & 74.10 & 74.69\\
        OFD~\cite{heo2019comprehensive}	     & 70.98 & 74.33 & 74.95 & 69.04 & 75.98 & 76.82\\
        CRD~\cite{tian2019contrastive}	     & 71.16 & 74.14 & 75.51 & 69.11 & 75.11 & 75.65\\
        ReviewKD~\cite{chen2021distilling} & 71.89 & 75.09 & 75.63 & 69.89 & 77.45 & 77.78\\
        \hline
        \multicolumn{7}{c}{\textit{Logit distillation methods}}\\
        KD~\cite{hinton2015distilling}	     & 70.66 & 73.54 & 73.33 & 67.65 & 74.07 & 74.45\\
        DIST~\cite{huang2022knowledge}     & 71.78 & 74.42 & 75.79 & 69.17 & 75.23 & 76.08\\
        DKD~\cite{zhao2022decoupled}      & 71.97 & 74.81 & 75.44 & 70.35 & 76.45 & 77.07\\
        \hline
        \textbf{KD++} & \textbf{72.53}(+1.87) & 74.59(+1.05) & 75.54(+2.21) & 70.10(+2.35) & 75.45(+1.38) & 76.42(+1.97)\\
        \textbf{DIST++} & 72.52(+0.74) & 75.00(+0.58) & 76.13(+0.34) & 69.80(+0.63) & 75.60(+0.37) & 76.64(+0.56)\\
        \textbf{DKD++} & 72.16(+0.19) & 75.02(+0.21) & \textbf{76.28}(+0.84) & \textbf{70.82}(+0.47) & 77.11(+0.66) & 77.49(+0.42)\\
        \textbf{ReviewKD++} & 72.05(+0.16) & \textbf{75.66}(+0.57) & 76.07(+0.44) & 70.45(+0.56) & \textbf{77.68}(+0.23) & \textbf{77.93}(+0.15)\\
        \hline
        \end{tabular}
    }
        \vspace{-1mm}
        \label{tab:cifar100-sota}
\end{table}

We conduct experiments to validate the proposed regularization techniques on feature norms and directions in the context of image classification and object detection.
Section~\ref{sec:setup} describes datasets and implementation details.
Section~\ref{sec:sota} benchmarks our approaches and existing KD methods.
Section~\ref{sec:ablation} ablates our losses with extensive analyses.

\subsection{Settings}\label{sec:setup}

For fair comparisons, our implementation adheres to the previous methodologies outlined in~\cite{tian2019contrastive, chen2021distilling, zhao2022decoupled, huang2022knowledge}. The hyperparameters $\alpha$ and $\beta$ are determined through an exhaustive search conducted within a predefined range, aligning with the established practices in prior studies.

\textbf{CIFAR-100}~\cite{krizhevsky2009learning} contains 50k training images and 10k testing images. For each input image, 4 pixels are added as padding on each side, and a $32\times32$ cropping patch is randomly selected from the padded images or their horizontally flipped counterparts. We employ weight initialization as described in~\cite{he2015delving}, training all student networks from scratch, while the teachers load the publicly available weights from~\cite{tian2019contrastive}. The student networks are trained using a mini-batch size of 128 over 240 epochs (with a linear warmup for the first 20 epochs), employing SGD with a weight decay of 5e-4 and momentum of 0.9. We set the initial learning rate of 0.1 for ResNet~\cite{he2016deep} and WRN~\cite{zagoruyko2016wide} backbones, and 0.02 for MobileNet~\cite{sandler2018mobilenetv2} and ShuffleNet~\cite{ma2018shufflenet} backbones, decaying it with a factor of 10 at 150th, 180th, and 210th. The temperature is empirically set to 4. 

\textbf{ImageNet}~\cite{russakovsky2015imagenet} comprises 1.28 million training images and 50,000 validation images spanning by 1,000 categories. We employ SGD with a mini-batch size of 512 for a total of 100 epochs (with a linear warmup for the first 5 epochs). The initial learning rate is set to 0.2 and is reduced by a factor of 10 every 30 epochs. Besides, the weight decay and momentum are set to 1e-4 and 0.9, respectively. The pre-trained weights for teachers come from PyTorch\footnote{https://pytorch.org/vision/stable/models.html} and TIMM~\cite{rw2019timm} for fair comparisons. The temperature for knowledge distillation is set to 1.

\textbf{COCO} 2017~\cite{lin2014microsoft} consists of 80 object categories with 118k training images and 5k validation images. We utilize Faster R-CNN~\cite{ren2015faster} with FPN~\cite{lin2017feature} as the feature extractor, wherein both teacher and student models adopt ResNet~\cite{he2016deep}. In addition, MobileNet-V2~\cite{sandler2018mobilenetv2} is used as a heterogeneous student model. All student models are trained with 1x scheduler, following Detectron2~\footnote{https://github.com/facebookresearch/detectron2}.


\subsection{Comparisons with State-of-the-art Results}\label{sec:sota}
\textbf{CIFAR-100 Classification}. Table~\ref{tab:cifar100-sota} showcases the performances of knowledge distillation on the CIFAR-100 dataset. In this context, spanning homogeneous and heterogeneous architectures, we undertake an extensive assessment over prominent \textit{feature distillation methods} (e.g., FitNet~\cite{romero2014fitnets}, RKD~\cite{park2019relational}, PKT~\cite{passalis2018probabilistic}, OFD~\cite{heo2019comprehensive}, CRD~\cite{tian2019contrastive}, ReviewKD~\cite{chen2021distilling}) and \textit{logits distillation methods}(e.g., KD~\cite{hinton2015distilling}, DIST~\cite{huang2022knowledge}, DKD~\cite{zhao2022decoupled}). The \textbf{++} signifies the integration of our novel ND loss into the preexisting methodologies.
A substantial conclusion can be derived from the Table~\ref{tab:cifar100-sota} that \textbf{our proposed ND loss manifests exceptional flexibility, which delivers advancements for both feature and logits distillation methodologies, irrespective of the homogeneity or heterogeneity for network architectures}. This phenomenon underscores the robust generalization prowess exhibited by the ND loss within the realm of knowledge distillation.


\begin{table}[t]
    \centering
    \caption{\small Benchmarking results on the ImageNet dataset. Methods are reported with top-1 accuracy (\%). ``T $\rightarrow$ S'' marks the architectures of {\tt teacher} and {\tt student}, short for knowledge distillation from the former to the latter.
    R\{18,34,50\} are the ResNet18, ResNet34, and ResNet50, respectively. MV1 means MobileNet-V1.
    Again, additionally using our ND loss, methods such as KD, ReviewKD, and DKD obtain better performance than their counterparts, achieving the state-of-the-art performance on this dataset.
    }
    \vspace{1mm}
        \renewcommand{\arraystretch}{1.2}
        \setlength\tabcolsep{1.5pt}
        \resizebox{\textwidth}{!}{%
        \begin{tabular}{ccc|ccc|cc|ccc}
        \hline
        T$\rightarrow $S & {\tt teacher} & {\tt student} & CRD~\cite{tian2019contrastive} & SRRL~\cite{yang2021knowledge} & ReviewKD~\cite{chen2021distilling} & KD~\cite{hinton2015distilling} & DKD~\cite{zhao2022decoupled} & \textbf{KD++} & \textbf{ReviewKD++} & \textbf{DKD++}\\
        \hline
        R34$\rightarrow $ R18 & 73.31 & 69.76 & 71.17 & 71.73 & 71.62 & 70.66 & 71.70 & 71.98 & 71.64 & \textbf{72.07}\\
        R50$\rightarrow $ MV1 & 76.16 & 68.87 & 71.37 & 72.49 & 72.56 & 70.50 & 72.05 & 72.77 & \textbf{72.96} & 72.63\\
        \hline
        \end{tabular}
    }
        \label{tab:imagenet-sota}
\end{table}

\textbf{ImageNet Classification}. We delve deeper into the efficacy of the proposed ND loss on the more expansive ImageNet dataset. Table~\ref{tab:imagenet-sota} provides supplementary evidence of the flexibility. Remarkably, despite its inherent simplicity, our \textbf{KD++} approach, which seamlessly integrates the ND loss into the naive \textbf{KD} framework, competes head-to-head with the SOTA results (\textbf{KD++} \textit{vs.} (\textbf{ReviewKD}, \textbf{DKD}) in Table~\ref{tab:cifar100-sota}\&\ref{tab:imagenet-sota}). Even, it surpasses the existing leading benchmarks on the extensive ImageNet dataset (Table~\ref{tab:imagenet-sota}), achieving notable improvements (\textbf{KD++}$_{R34\rightarrow R18}$: 71.98\% \textit{vs.} \textbf{SRRL}$_{R34\rightarrow R18}$: 71.73\%, \textbf{KD++}$_{R50\rightarrow MV1}$: 72.77\% \textit{vs.} \textbf{ReviewKD}$_{R50\rightarrow MV1}$: 72.56\%).

\textbf{COCO Object Detection}. We verify the efficacy of the proposed ND loss in knowledge distillation tasks for object detection on the COCO dataset, as shown in Table~\ref{tab:coco}. Specifically, the \textbf{ReviewKD++} yields a significant improvement in performance, outperforming state-of-the-art results with a remarkable margin.

\begin{table}[t]
    \centering
    \small
        \caption{\small Detection results (mAP in $\%$) on the \textbf{COCO \texttt{val2017}} using Faster R-CNN detector.
        Incorporating our ND loss, KD++ and ReviewKD++ obtain performance gains over their original counterparts,
        achieving the state-of-the-art KD performance.
        }
        \renewcommand{\arraystretch}{1.1}
        \resizebox{0.95\textwidth}{!}{
        \small
        \begin{tabular}{c|ccc|ccc|ccc}
        \hline
        ~ & \multicolumn{3}{c}{R101$\rightarrow$R18} & \multicolumn{3}{|c}{R101$\rightarrow$R50} & \multicolumn{3}{|c}{R50$\rightarrow$MV2}\\
        Methods & mAP & AP$^{50}$ & AP$^{75}$ & mAP & AP$^{50}$ & AP$^{75}$ & mAP & AP$^{50}$ & AP$^{75}$\\
        \hline
        {\tt teacher} & 42.04 & 62.48 & 45.88 & 42.04 & 62.48 & 45.88 & 40.22 & 61.02 & 43.81\\
        {\tt student} & 33.26 & 53.61 & 35.26 & 37.93 & 58.84 & 41.05 & 29.47 & 48.87 & 30.90\\
        \hline
        KD~\cite{hinton2015distilling}     & 33.97 & 54.66 & 36.62 & 38.35 & 59.41 & 41.71 & 30.13 & 50.28 & 31.35\\
        FitNet\cite{romero2014fitnets}                             & 34.13 & 54.16 & 36.71 & 38.76 & 59.62 & 41.80 & 30.20 & 49.80 & 31.69\\
        FGFI\cite{wang2019distilling}                              & 35.44 & 55.51 & 38.17 & 39.44 & 60.27 & 43.04 & 31.16 & 50.68 & 32.92\\
        DKD~\cite{zhao2022decoupled}       & 35.05 & 56.60 & 37.54 & 39.25 & 60.90 & 42.73 & 32.34 & 53.77 & 34.01\\
        ReviewKD~\cite{chen2021distilling} & 36.75 & 56.72 & 34.00 & 40.36 & 60.97 & 44.08 & 33.71 & 53.15 & 36.13\\
        \hline
        \textbf{KD++} & 36.12 & 56.81 & 37.64 & 39.86 & 61.07 & 43.57 & 33.26 & 53.71 & 34.85\\
        \textbf{ReviewKD++} & \textbf{37.43} & \textbf{57.96} & \textbf{40.15} & \textbf{41.03} & \textbf{61.80} & \textbf{44.94} & \textbf{34.51} & \textbf{55.18} & \textbf{37.21}\\
        \hline
    \end{tabular}
    }
        \label{tab:coco}
\end{table}

\subsection{Ablation Study}\label{sec:ablation}

In this subsection, we first investigate the ablation experiments pertaining to feature norm and direction regularization. Subsequently, we offer a visual analysis of the impact of ND before and after applying to CIFAR-10. Finally, we conduct intriguing experiments on ImageNet and observe that our approach accrues advantages from employing larger teacher models.

\textbf{The isolation of feature norm and direction regularization.} 
Recall that Section~\ref{sec:FNR} and Section~\ref{sec:FDR} explore the concrete instantiation of feature norm and direction regularization separately. Owing to space limitations, we present only simple test results for $\mathcal{L}_2$ (Eq.~\ref{eq:MSE}) and SIFN (Eq.~\ref{eq:SFN}) on CIFAR-100 in Table~\ref{tab:only_norm}. Yet additional offline experiments substantiate that SIFN outperforms $\mathcal{L}_2$ regularization in terms of performance and consistently affirm that large student norms encapsulate more teacher knowledge. Similarly, Table~\ref{tab:only_direction} demonstrates the superior gains of cosine (Eq.~\ref{eq:cosine}) compared to InfoNCE (Eq.~\ref{eq:nce}), further underscoring the significance of feature direction constraints.

\begin{table}[b]
\caption{\small
\textbf{Analysis of feature norm and direction regularization.} 
We train {\tt teacher} (ResNet-50) and {\tt student} (MobileNet-V2) models on the CIFAR100 dataset and report accuracy (\%) on its test-set.
We use KD~\cite{hinton2015distilling} as the \emph{baseline}, which is a logit distillation method.
From (a-b), we see that applying either norm or direction regularization on {\tt student} features improve KD as shown by the increased {\tt student} accuracy.
While combining both outperforms \emph{baseline} (c), using ND loss achieves the best (d).
}
    \begin{subtable}[h]{0.18\textwidth}
        \renewcommand{\arraystretch}{1.2}
        \setlength\tabcolsep{1.5pt}
        \small
        \centering
        \caption{Regularizing feature norm only.}
        \vspace{-2mm}
        \begin{tabular}{lc}
        case &  acc.\\
        \hline
        \toprule
        \textit{baseline} & 67.65\\
        $\mathcal{L}_2$ & 69.05\\
        SIFN & 69.32\\
         &\\
        \end{tabular}
        \label{tab:only_norm}
    \end{subtable}
    \hfill
    \begin{subtable}[h]{0.28\textwidth}
        \renewcommand{\arraystretch}{1.2}
        \setlength\tabcolsep{1.5pt}
        \small
        \centering
        \caption{Regularizing feature direction only.}
        \vspace{-2mm}
        \begin{tabular}{lcc}
        case & R50-MV2 & R56-R20\\
        \hline
        \toprule
        \textit{baseline} & 67.65 & 70.66\\
        cosine & 69.18 & 71.75\\
        InfoNCE   & 69.06 & 70.73\\
        &\\
        \end{tabular}
        \label{tab:only_direction}
    \end{subtable}
    \hfill
    \begin{subtable}[h]{0.25\textwidth}
        \renewcommand{\arraystretch}{1.2}
        \setlength\tabcolsep{1.5pt}
        \small
        \centering
        \caption{Regularizing both feature norm and direction.}
        \vspace{-2mm}
        \begin{tabular}{lc}
        case &  acc. \\
        \hline
        \toprule
        cosine + $\mathcal{L}_2$ & 68.62\\
        cosine + SIFN & 69.07\\
        InfoNCE + $\mathcal{L}_2$ & 68.47\\
        InfoNCE + SIFN & 68.71\\
        \end{tabular}
        \label{tab:dir+norm}
    \end{subtable}
    \begin{subtable}[h]{0.25\textwidth}
        \renewcommand{\arraystretch}{1.2}
        \setlength\tabcolsep{1.5pt}
        \small
        \centering
        \caption{The proposed ND loss works the best.}
        \vspace{-2mm}
        \begin{tabular}{lc}
        case &  acc. \\
        \hline
        \toprule
        CE + KL (\emph{baseline}) & 67.65\\
        CE + ND & 68.78\\
        KL + ND & 68.68\\
        CE + KL + ND  & \textbf{70.10}\\
       \end{tabular}
       \label{tab:EFA}
    \end{subtable}
\label{tab:ablation}
\end{table}

\textbf{ND loss yields better results.} This part discusses the benefits of the independent amalgamation of feature norm and direction regularization. Table~\ref{tab:dir+norm} consolidates feature direction regularization (cosine, InfoNCE) and feature norm ($\mathcal{L}_2$, SIFN), unveiling that the optimal setting (cosine + SIFN) leads to superior performance (69.07\%) among all combinations. Nevertheless, upon meticulous scrutiny, it becomes apparent that directly integrating feature direction with norm regularization can prove deleterious, as it engenders lower results than separate regularization. For instance, (cosine + $\mathcal{L}_2$) or (cosine + SIFN) reduces accuracy from 69.18\% to 68.62\% (-0.56\%) and 69.07\% (-0.11\%), respectively. Similarly, (SIFN + cosine) or (SIFN + InfoNCE) results in a substantial decline from 69.32\% to 69.07\% (-0.25\%) and 68.71\% (-0.61\%), respectively. In contrast, the proposed ND loss capitalizes on the merits of both strategies, culminating in remarkable achievements of 70.10\% (Table~\ref{tab:EFA}).

\begin{figure}[t]
\centering
    \subcaptionbox{{\tt teacher}}
    {\includegraphics[width = 0.24\textwidth, height=3.1cm]{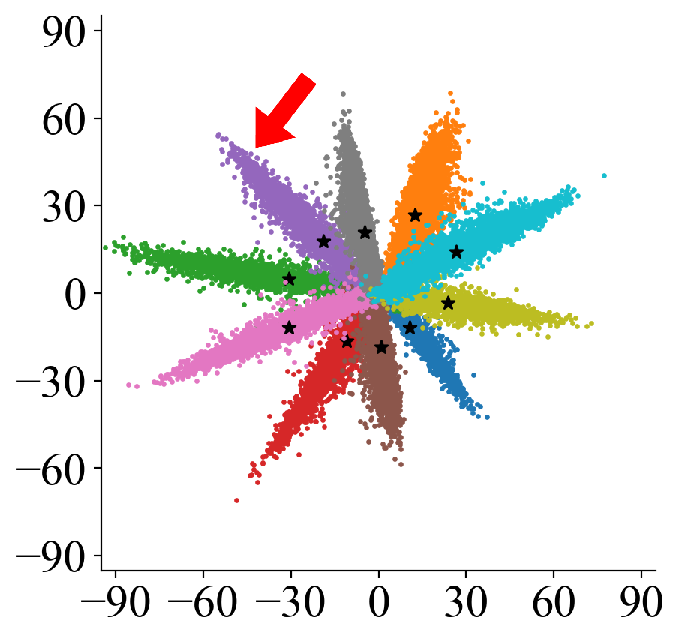}}
    \hfill
    \subcaptionbox{naive small model}
    {\includegraphics[width = 0.24\textwidth, height=3.1cm]{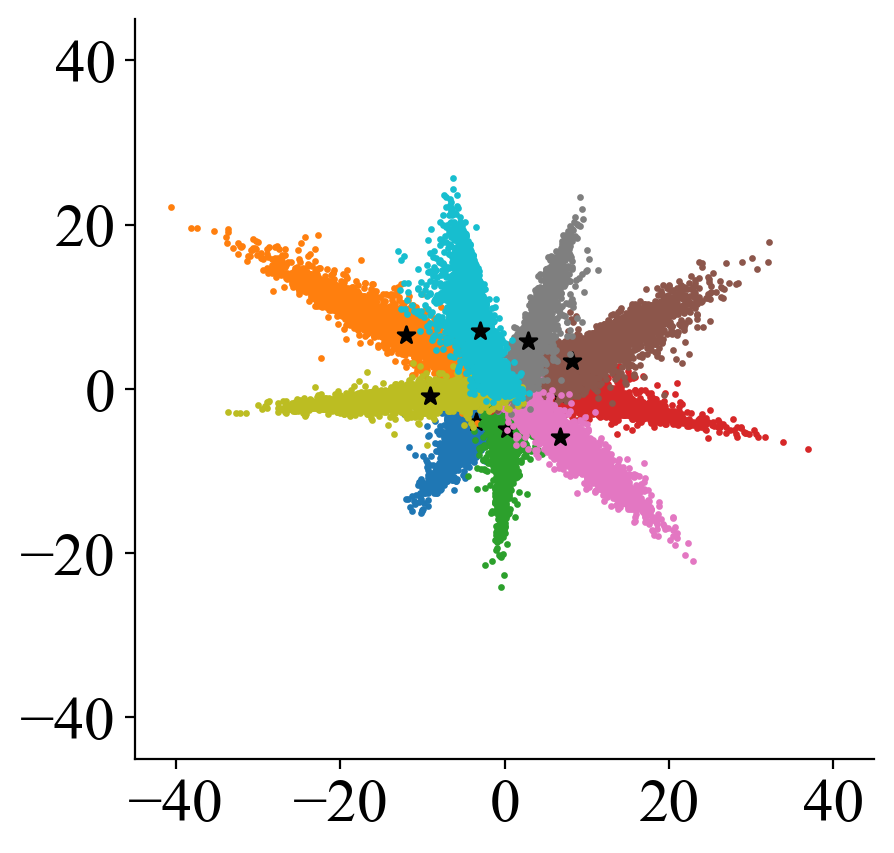}}
    \hfill
    \subcaptionbox{{\tt student} by KD}
    {\includegraphics[width = 0.24\textwidth, height=3.1cm]{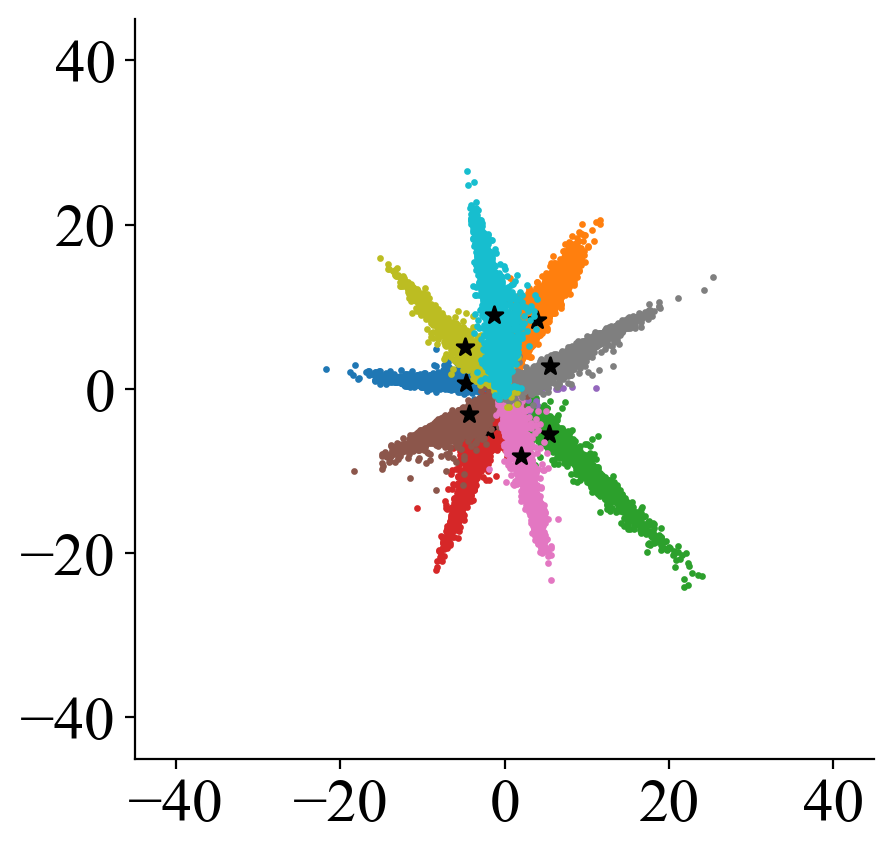}}
    \hfill
    \subcaptionbox{{\tt student} by KD++}
    {\includegraphics[width = 0.24\textwidth, height=3.1cm]{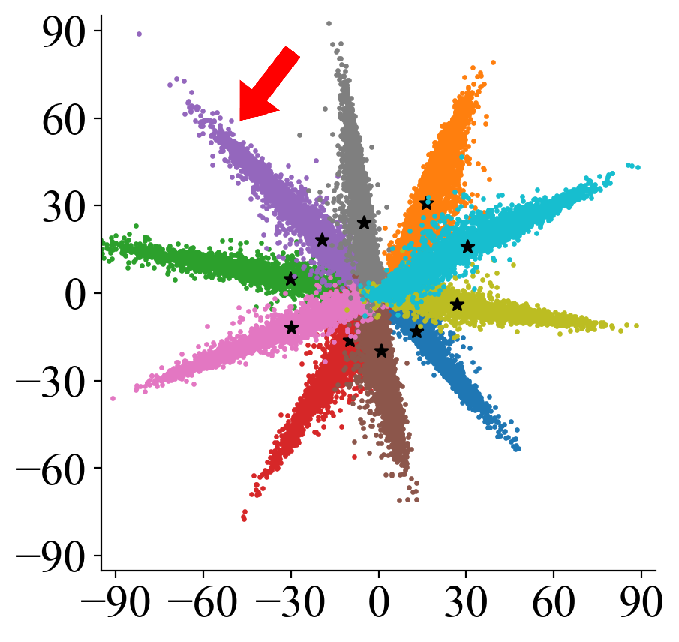}}
\vspace{-1mm}
\caption{\small \textbf{Visualization of 2D embedding features.} 
(a) Features computed by {\tt teacher} (ResNet-50) are well separated at class label; note the \textcolor{violet}{purple} class pointed by \textcolor{red}{red arrow}.
(b) a small-capacity model (ResNet-18) fails to separate this class, which is occluded by others.
(c) Even using KD~\cite{hinton2015distilling} to train ResNet-18 {\tt student} cannot reveal this \textcolor{violet}{purple} class.
(d) Using our ND loss along with KD, i.e., KD++, achieves better separation of the points and reveals \textcolor{violet}{purple} class.
This attributes to the feature direction regularization using {\tt teacher} class-means. 
Moreover, {\tt student} features in (d) have larger-norms than the {\tt teacher} in (a).
}
\vspace{-2mm}
\label{fig:ablation-res50-18} 
\end{figure}

\textbf{KD++ as a stronger baseline.}
Table~\ref{tab:EFA} illustrates the impacts of different losses in canonical knowledge distillation. By incorporating ND loss into conventional KD framework~\cite{hinton2015distilling}, \textbf{KD++} (i.e., CE+KL+ND)  achieves a stunning result (\textbf{KD} (67.65\%)$\rightarrow$ \textbf{KD++} (70.10\%)). Interestingly, combining ND alone with CE or KL can also boost the accuracy by about $1\%$ compared to classical KD. It is worth noting that \textbf{KD++} introduces virtually no additional parameters and minimal computational overhead, making it a stronger baseline for knowledge distillation (more validations can be gleaned from the results presented in Table~\ref{tab:cifar100-sota}\&\ref{tab:imagenet-sota}\&\ref{tab:coco}). In addition, we visually examine the feature with a learnable dimension reduction approach~\cite{wen2016discriminative}, as shown in Fig.~\ref{fig:ablation-res50-18}. First, as indicated in Fig.~\ref{fig:ablation-res50-18}d, \textbf{KD++} demonstrates notably amplified feature norms, surpassing even those of the teacher depicted in Fig.~\ref{fig:ablation-res50-18}a. Furthermore, the directions in \textbf{KD++} align well with the teacher (Fig.~\ref{fig:ablation-res50-18}a \textit{vs.}  Fig.~\ref{fig:ablation-res50-18}d, thereby maintaining consistent relative margins among categories. Another observation is that both the original student model (Fig.~\ref{fig:ablation-res50-18}c) and the conventional KD (Fig.~\ref{fig:ablation-res50-18}c) exhibit direct failures in classifying the \textcolor{purple}{purple} category, whereas our approach, \textbf{KD++} (Fig.~\ref{fig:ablation-res50-18}d), effectively reattends to the "disappeared" category. 

\begin{table}[t]
    \centering
            \caption{\small \textbf{Our method could benefit from larger teachers.} Methods are reported with top-1 accuracy (\%) on the ImageNet validation set. 
            With {\tt teacher} capacity increasing, {\tt student} models (trained with our ND loss)  achieve better classification results.
            Yet, previous KD methods do not necessarily obtain better results by distilling larger teachers.
            $*$ represents our implementation based on the official code.
            }        
        \renewcommand{\arraystretch}{1.2}
        \setlength\tabcolsep{3.5pt}
        \resizebox{\textwidth}{!}{%
        \begin{tabular}{cccc|ccc|ccc}
        \hline
        {\tt student} & {\tt teacher} & {\tt student} & {\tt teacher} & KD$^*$\cite{hinton2015distilling} & ReviewKD$^{*}$\cite{chen2021distilling} & DKD$^{*}$\cite{zhao2022decoupled} & \textbf{KD++} & \textbf{ReviewKD++} & \textbf{DKD++}\\
        \hline
        \multirow{4}{*}{ResNet-18} & ResNet-34 & \multirow{4}{*}{69.76} & 73.31 & 70.66 & 71.62 & 71.70 & 71.98 & 71.64 & \textbf{72.07}\\
        ~ & ResNet-50  & ~ & 76.16 & 71.35 & 71.10 & 71.87 & \textbf{72.53} & 71.71 & 72.08\\
        ~ & ResNet-101 & ~ & 77.37 & 71.09 & 70.98 & 72.10 & \textbf{72.54} & 71.77 & 72.26\\
        ~ & ResNet-152 & ~ & 78.31 & 71.12 & 71.36 & 71.97 & \textbf{72.54} & 71.79 & 72.48\\
        \hline
        \multirow{2}{*}{ResNet-18} & ViT-S & \multirow{2}{*}{69.76} & 74.64 & 71.32 & n/a & 71.21 & \textbf{71.46} & n/a & 71.33\\
        ~ & ViT-B & ~ & 78.00 & 71.63 & n/a & 71.62 & \textbf{71.84} & n/a & 71.69\\
        \hline
        \end{tabular}
    }
        \label{tab:imagenet-largetea}
\end{table}
\textbf{Benefit from larger teacher models.}
Since previous experiments highlight that consistent direction with a larger norm for student models can better facilitate the assimilation of knowledge from teacher models, we further investigate whether our approach exhibits monotonic incremental gains when faced with larger teacher models. As depicted in Table~\ref{tab:imagenet-largetea}, it is evident that for KD, ReviewKD, and DKD show a degradation or fluctuation trend when scaling up the teacher models from ResNet34 to ResNet152. However, upon incorporating our ND loss, the results showcase a consistent improvement (e.g., \textbf{DKD++}: 72.07\% $\rightarrow$ 72.08\% $\rightarrow$ 72.26\% $\rightarrow$ 72.48\%) or reaching a saturation point (e.g., \textbf{KD++}: 71.98\% $\rightarrow$ 72.53\% $\rightarrow$ 72.54\% $\rightarrow$ 72.54\%). Besides, we extend our experiments to include distillation from Transformer~\cite{dosovitskiy2020image} to ResNet, further reinforcing this observation. Nonetheless, owing to the architectural differences, specifically the contrasting characteristics of global attention in Transformer and local receptive fields in Convolution, the benefits are not as conspicuous as in cases with homogeneous architectures.



\section{Discussion and Conclusion}

{\bf Broader Impacts and Limitations.}
As our work falls in the area of knowledge distillation, 
we do not see any new potential societal impacts other than those already known, e.g., {\tt student} models might learn bias and unfairness delievered by the {\tt teacher}.
Our work has some visible limitations, e.g., we apply ND loss to the penultimate layer only, and we do not study how to distill large pretrained models (e.g., language models).
Addressing these are important and  future work.

{\bf Conclusion.}
We study feature regularization w.r.t norm and direction when training {\tt student} models for better knowledge distillation (KD).
Indeed, experiments demonstrate that doing so with our explored simple methods and the proposed ND loss help existing KD methods achieve better performance.
We expect the proposed ND loss to be a plug-in in future KD methods.

\bibliographystyle{unsrt}
\bibliography{references}

\clearpage
\appendix
\newpage

\section{More Implementation Details}
For fair comparisons, we choose network architectures from recent studies~\cite{heo2019comprehensive, tian2019contrastive, chen2021distilling, zhao2022decoupled, huang2022knowledge}. We experiment with various network architectures on CIFAR dataset, including ResNet~\cite{he2016deep}, WideResNet~\cite{zagoruyko2016wide}, MobileNet~\cite{sandler2018mobilenetv2}, and ShuffleNet~\cite{Zhang_2018_CVPR, ma2018shufflenet}.
For homogeneous knowledge distillation challenges, we consider ResNet-56$ \rightarrow $ResNet-20, WRN-40-2 $ \rightarrow $ WRN-40-1, ResNet-32$\times$4 $ \rightarrow $ ResNet-8$\times$4 as the {\tt teacher} $ \rightarrow $ {\tt student} models on the CIFAR dataset, which is characterized by a relatively low input resolution (32$\times 32$). ResNet-20 and ResNet-56 comprise three fundamental blocks with channel dimensions 16, 32, and 64. WRN-d-r denotes a WideResNet backbone with a depth of \textit{d} and a width factor of \textit{r}. ResNet-8$\times$4 and ResNet-32$\times$4 define networks with (16, 32, 64)×4 channels, corresponding to 64, 128, and 256 channels, respectively. In the context of heterogeneous knowledge distillation, we opted for MobileNet and ShuffleNet as the student models, adhering to the configuration of previous studies~\cite{tian2019contrastive, chen2021distilling}. To adapt MobileNetV2 specifically to the CIFAR dataset, we introduce a widening factor of 0.5. For the larger-scale ImageNet dataset, we employ the standard ResNet, the MobileNet, and the typical Transformer~\cite{dosovitskiy2020image} backbones, consistent with previous approaches~\cite{heo2019comprehensive, tian2019contrastive, chen2021distilling, zhao2022decoupled, huang2022knowledge}.

Our proposed ND loss function regularizes the norm and direction of the {\tt student} features at the penultimate layer before logits. The embedding features of the  {\tt student} and {\tt teacher} models may have different dimensions. This can be addressed by learning a fully connected layer (followed by Batch Normalization) with the {\tt student} to project its features to the same dimension as the {\tt teacher}'s.

\begin{figure}[h]
\centering
    \subcaptionbox{sensitivity of $\beta$, (set $\alpha=1$)}
    {\includegraphics[width=0.4\textwidth]{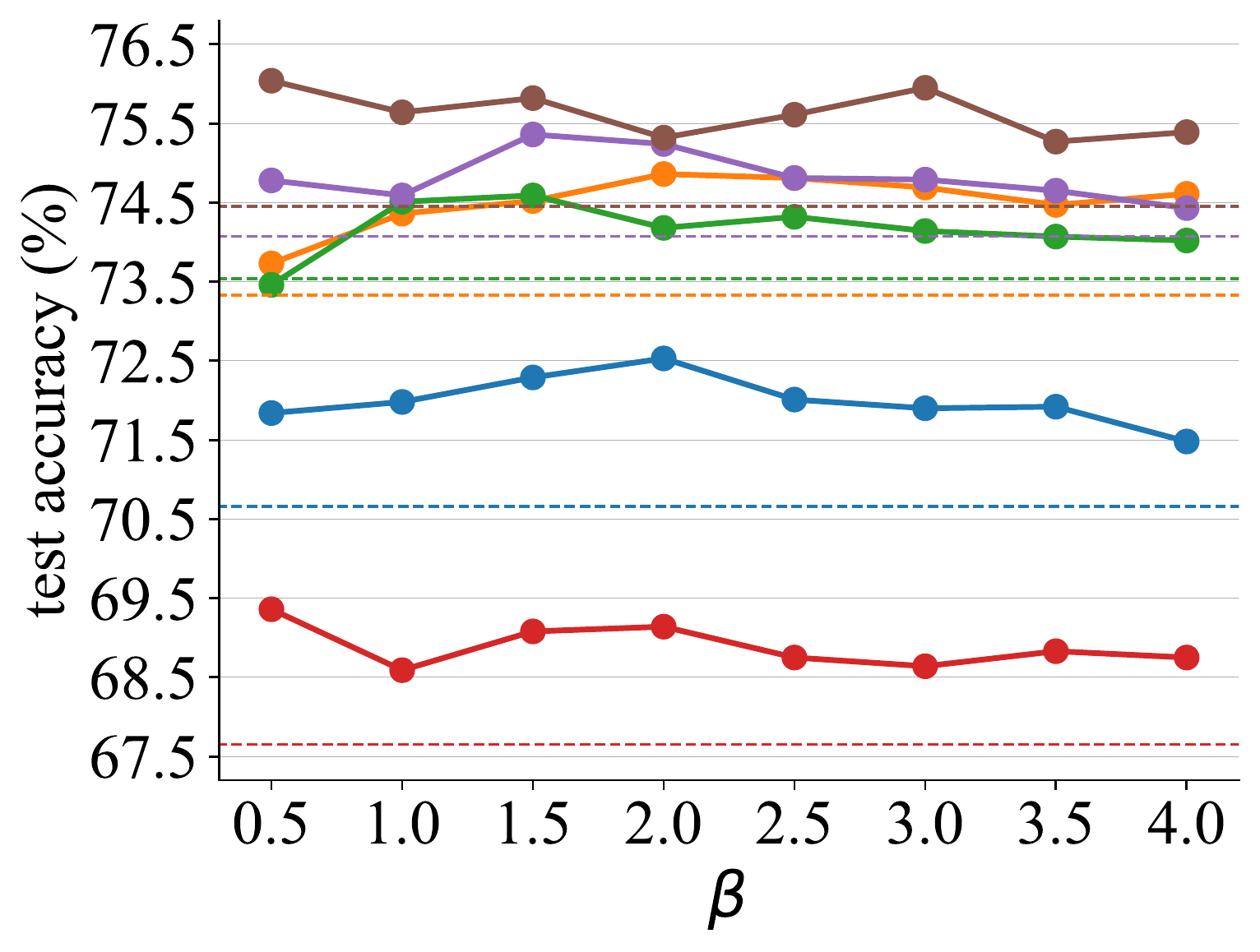}}
    \hfill
    \subcaptionbox{sensitivity of $\alpha$, (select the best $\beta$)}
    {\includegraphics[width=0.4\textwidth,height=4.2cm]{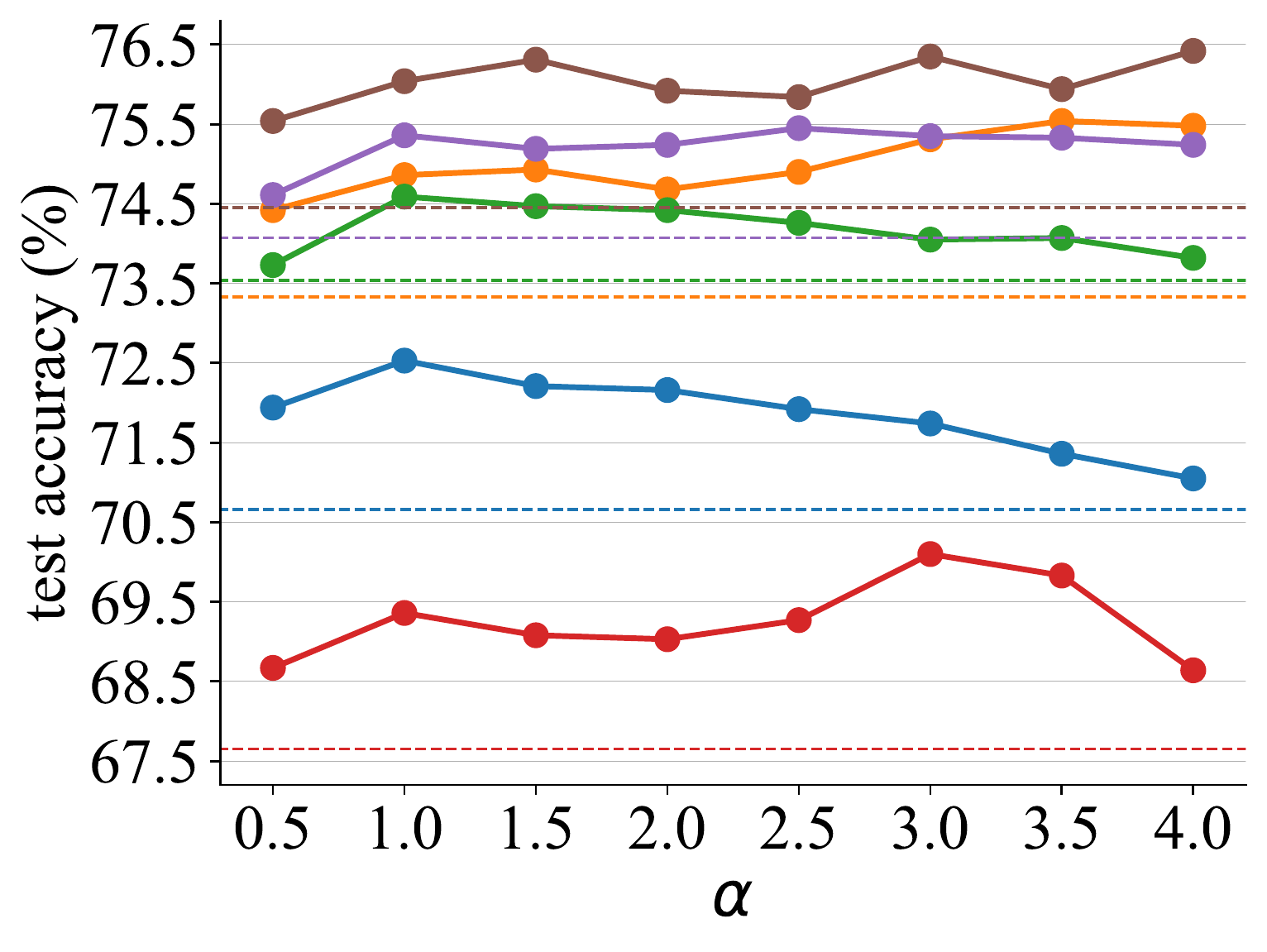}}
    \hfill
    \includegraphics[width=0.18\textwidth,height=4.2cm]{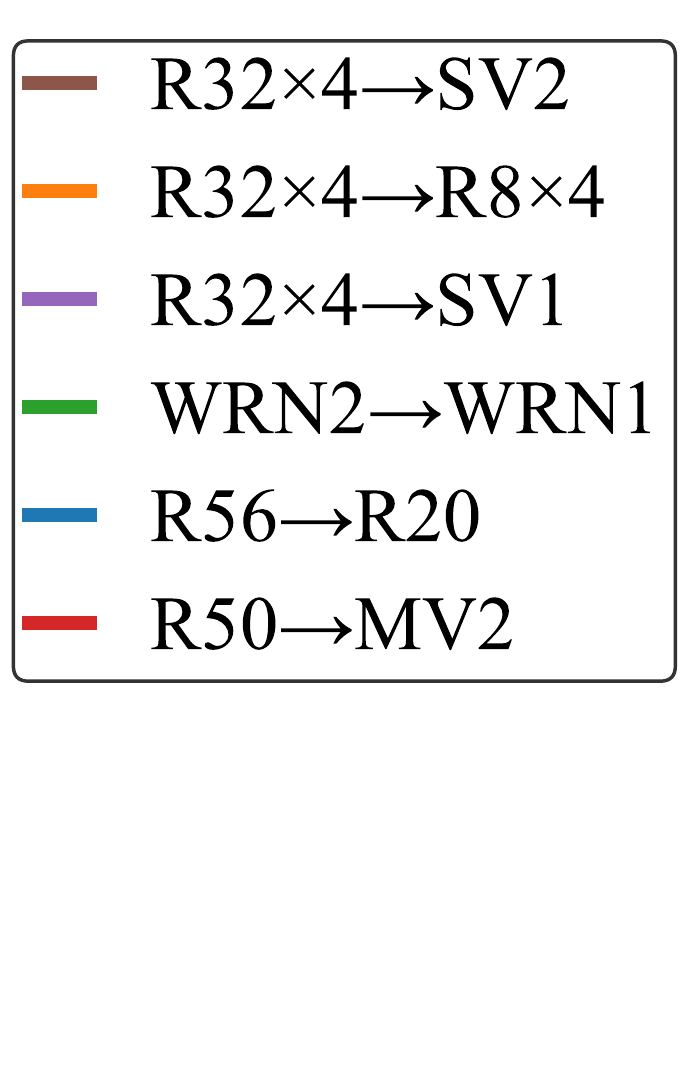}

\caption{\small
The impact of hyper-parameters $\alpha$ and $\beta$. The dashed lines illustrate the performance based on standard KD loss(corresponding to the specific setting($\alpha$ = 1 and $\beta$ = 0)). (a). $\alpha$ is set to 1, then evaluate the impact of $\beta$. (b). keep best $\beta$ fixed, assessing the impact of $\alpha$.
}
\vspace{-2.0mm}
\label{fig:suppl_alpha_beta}
\end{figure}

\section{Additional Ablation Studies}

\subsection{The Impact of Hyper-parameters $\alpha$ and $\beta$}
In the Eq.9, we introduce the KD++ loss function as $\mathcal{L} = \mathcal{L}_{ce} + \alpha\mathcal{L}_{kd} + \beta\mathcal{L}_{nd}$. As elucidated in the experiment details, the values of $\alpha$ and $\beta$ are acquired through an exhaustive search within a predefined range. To substantiate the efficacy of the proposed ND loss, we conduct extensive experiments aiming at probing the sensitivities of the hyperparameters $\alpha$ and $\beta$, as depicted in Fig.~\ref{fig:suppl_alpha_beta}. The dashed lines illustrate the standard KD loss(corresponding to specific setting($\alpha$ = 1 and $\beta$ = 0)) in Fig.~\ref{fig:suppl_alpha_beta}a. Evidently, our proposed ND loss consistently surpasses the scenario devoid of ND loss as $\beta$ ranges from 0.5 to 4.0 (the solid line always surpasses the dashed line for the same color). Furthermore, in Fig.~\ref{fig:suppl_alpha_beta}b, when the optimal $\beta$ value is fixed, the distilled performance exhibits consistent enhancement compared to the baseline as $\alpha$ varies. These results compellingly attest to the overarching efficacy of the proposed ND loss in our experiments, with the sensitivity of hyperparameters merely influencing the magnitude of improvement. 

Certainly, an alternative approach worth contemplating for acquiring optimal parameters entails performing a grid search within the hyperplane spanning by $\alpha$ and $\beta$. Nevertheless, such an approach incurs heightened intricacy and computational demands. The goal of this study, however, resides in substantiating the efficacy of the proposed ND loss, thereby necessitating the demonstration that outcomes attained with non-zero $\beta$ surpass those achieved through the conventional KD setting ($\alpha$=1 and $\beta$=0). In practical scenarios pertaining to knowledge distillation tasks, it becomes feasible to ascertain the optimal $\alpha$ and $\beta$ parameter pairs by undertaking a grid search across the $\alpha-\beta$ parameter space, while judiciously considering the facet of actual performance augmentation.

\begin{figure}[h]
\centering
    \resizebox{0.93\textwidth}{!}{%
    \subcaptionbox{ResNet-34$\rightarrow $ResNet-18}
    {\includegraphics[width=0.42\textwidth]{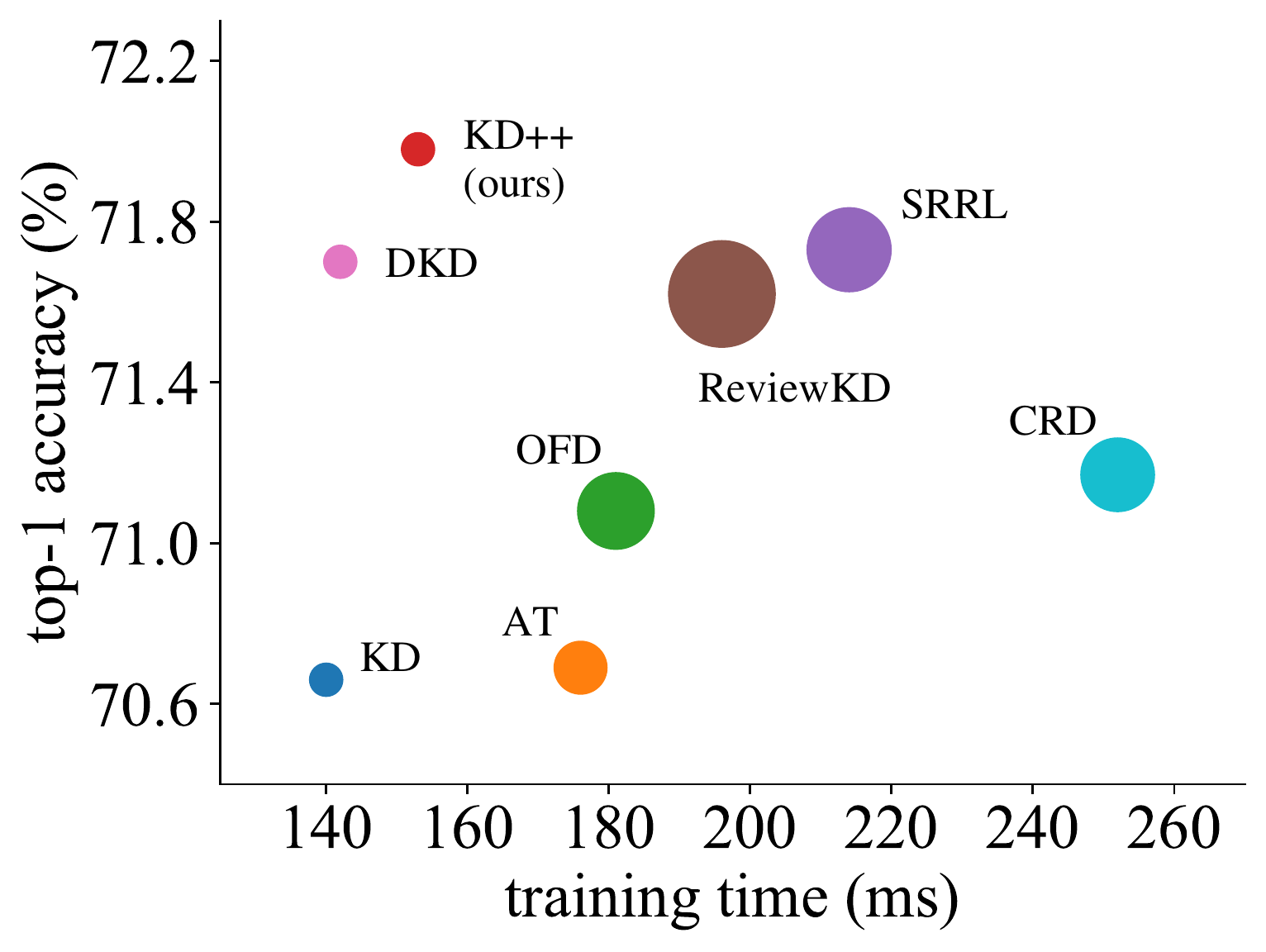}}
    \hfill
    \subcaptionbox{ResNet-50$\rightarrow $MobileNet-V1}
    {\includegraphics[width=0.42\textwidth]{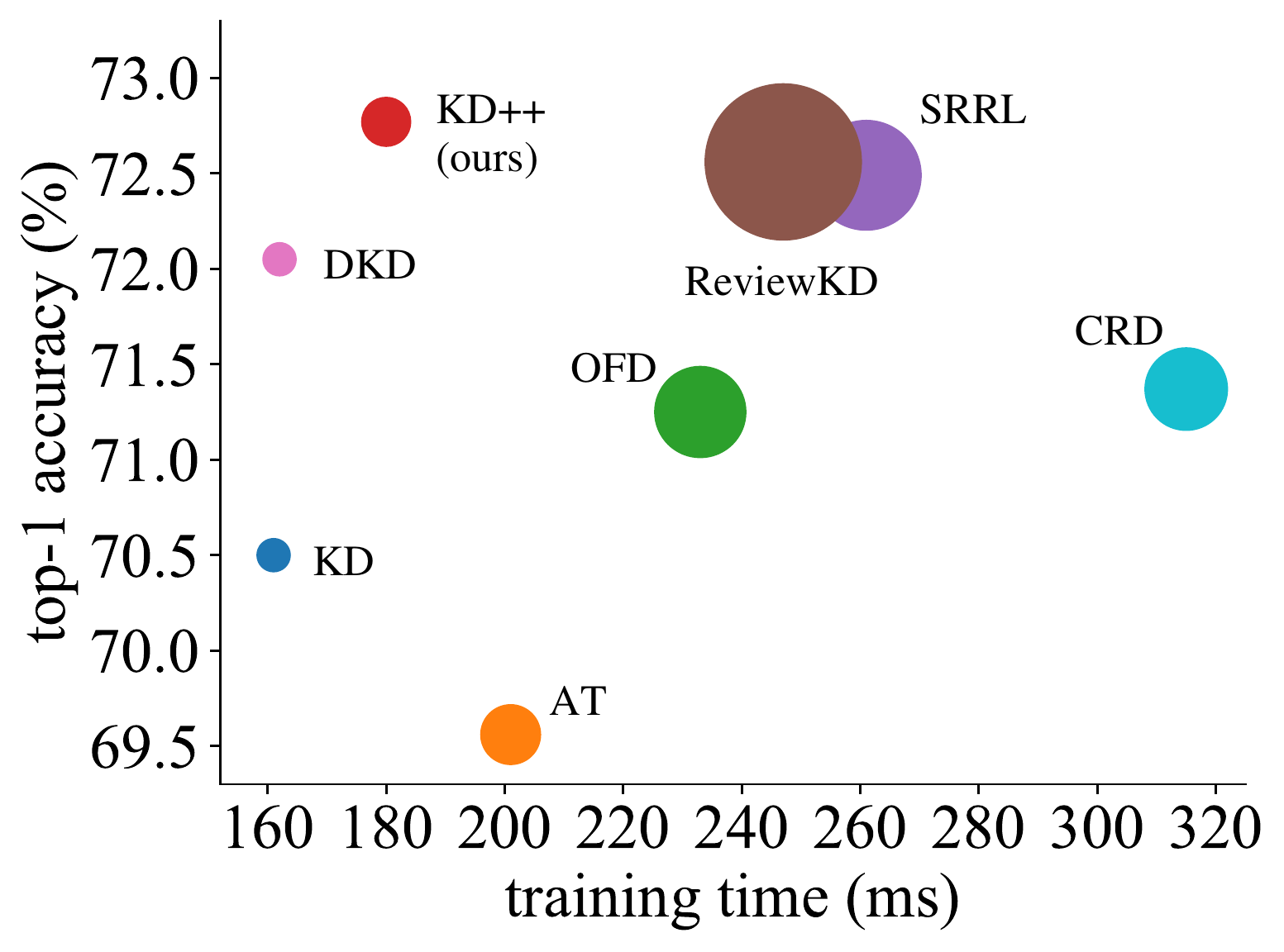}}
    }

\caption{\small
\textbf{Wall-clock time per training iteration vs. accuracy} on the ImageNet validation set. left: homogeneous architectures, right: heterogeneous architectures. Enlarged circles correspond to a higher demand for parameters.
}
\vspace{-2.0mm}
\label{fig:suppl_params_acc}
\end{figure}
\subsection{Complexity Comparisons}
In this subsection, we present simple comparisons for mainstream knowledge distillation methods, as illustrated in Fig.~\ref{fig:suppl_params_acc}. Fig.~\ref{fig:suppl_params_acc}a and Fig.~\ref{fig:suppl_params_acc}b showcase examples of homogeneous distillation (ResNet-34 $\rightarrow$ ResNet-18) and heterogeneous distillation (ResNet-50 $\rightarrow$ MobileNet-V1) on the ImageNet dataset. We measure the average time cost per batch iteration over the entire dataset as the horizontal axis and the Top-1 accuracy as the vertical axis. The varying sizes of circular markers representing different methods are proportional to the actual model parameter sizes. It is clear that our approach (KD++) delivers better performance with a small amount of time expense. It is important to highlight that in heterogeneous knowledge distillation tasks, there is typically a disparity in feature dimensions. Consequently, the inclusion of a bridging linear dimension transformation layer becomes imperative, attributing to the marginal increment in parameterization observed in our method, KD++, as compared to the classical KD approach.

\begin{table}[h]
    \centering
    \caption{\small Altering the norm of the {\tt teacher} mode with a scaling factor $m$. Classification accuracy on the CIFAR-100 test set. The gray background indicates the default setting.
    }
    \vspace{1mm}
        \renewcommand{\arraystretch}{1.2}
        \resizebox{0.85\textwidth}{!}{%
        \begin{tabular}{c|c| c| >{\columncolor{lightgray}} c|c|c|c|c|c|c}
        \hline
        $m$            & -0.5  & -0.1  & 0.0  & 0.1   & 0.5   & 0.7   & 1.0   & 1.5   & 2.0\\
        \hline
        R56$\rightarrow $ R20 & 71.57 & 72.19 & \textbf{72.53} & 71.76 & 71.86 & 71.64 & 71.79 & 71.74 & 71.92\\
        R50$\rightarrow $ MV2 & 69.46 & 69.43 & 70.10 & 70.17 & \textbf{70.23} & 69.68 & 69.72 & 68.49 & 69.44\\
        \hline
        \end{tabular}
    }
        \label{tab:suppl_tea_norm}
\end{table}
\subsection{Does the Magnitude of {\tt Teacher} Norm Matter ?}

In earlier sections, we discover that improving the student model's norm benefits knowledge distillation. Therefore, a natural question arises: does increasing the teacher model norm also contribute to improving student performance?  To investigate this, we conduct simple experiments where we introduce a scaling factor, denoted as $m$, to the norm of the teacher model in Eq.8 as follows:
\begin{equation}\label{eq:tea-norm}
    \mathcal{L}_{nd} 
= - \frac{1}{C}\sum_{k=1}^C \frac{1}{\vert{\cal I}_k\vert} 
    \sum_{i \in {\cal I}_k} \frac{\mathbf{f}^s_i \cdot \mathbf{e}_k}
{\max \left\{ \lVert \mathbf{f}^s_i \rVert_2,  \lVert \mathbf{f}^t_i \rVert_2 \cdot (1+m) \right\}}
\end{equation}

Interestingly, our experimental results indicate that in the context of homogeneous knowledge distillation, altering the norm of the teacher model, whether increasing or decreasing it, does not lead to better improvement in student performance compared to maintaining the original norm of the teacher model. However, in the case of heterogeneous knowledge distillation, there may be benefits in appropriately increasing the norm of the teacher features. It is worth noting that since this experiment has not been tested on a large-scale dataset, we cannot definitively conclude whether a larger teacher norm will always result in improvements. Nonetheless, this presents a promising direction for future exploration, where joint constraints on the norm size and direction can be applied to both teacher and student models.
\begin{figure}[h]
\centering
    \subcaptionbox{{\tt teacher} (Res56)}
    {\includegraphics[width = 0.24\textwidth, height=3.1cm]{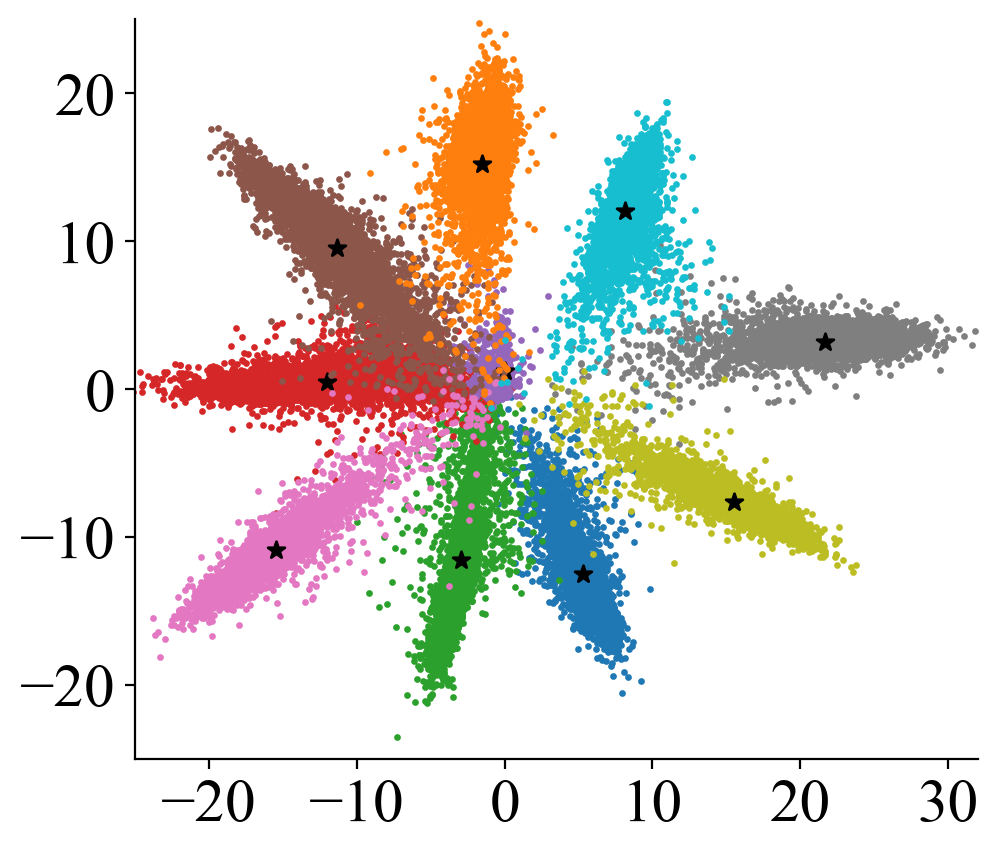}}
    \hfill
    \subcaptionbox{naive model (Res20)}
    {\includegraphics[width = 0.24\textwidth, height=3.1cm]{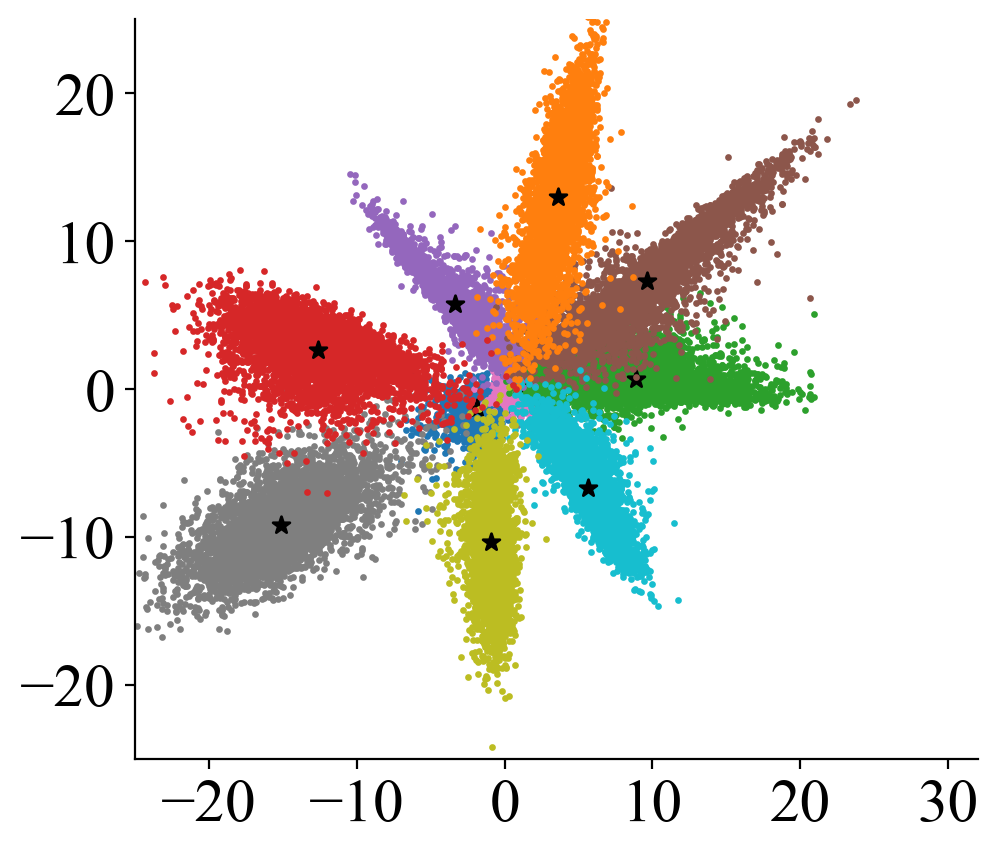}}
    \hfill
    \subcaptionbox{{\tt student} by KD}
    {\includegraphics[width = 0.24\textwidth, height=3.1cm]{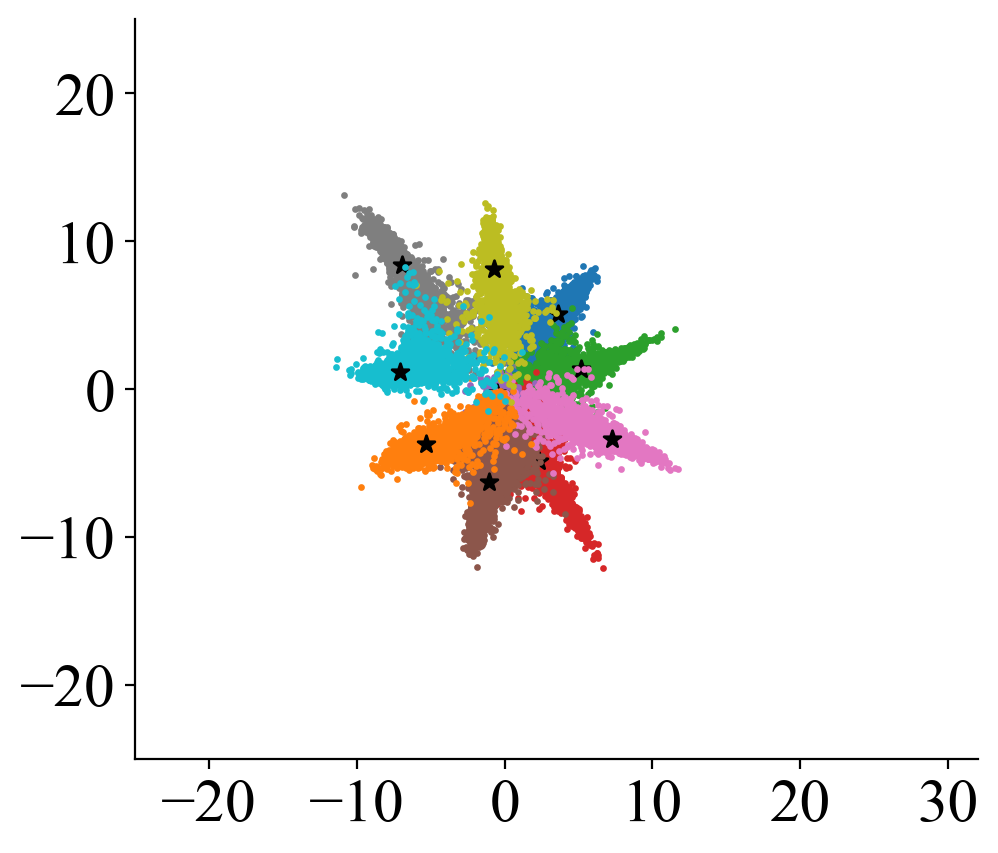}}
    \hfill
    \subcaptionbox{{\tt student} by KD++}
    {\includegraphics[width = 0.24\textwidth, height=3.1cm]{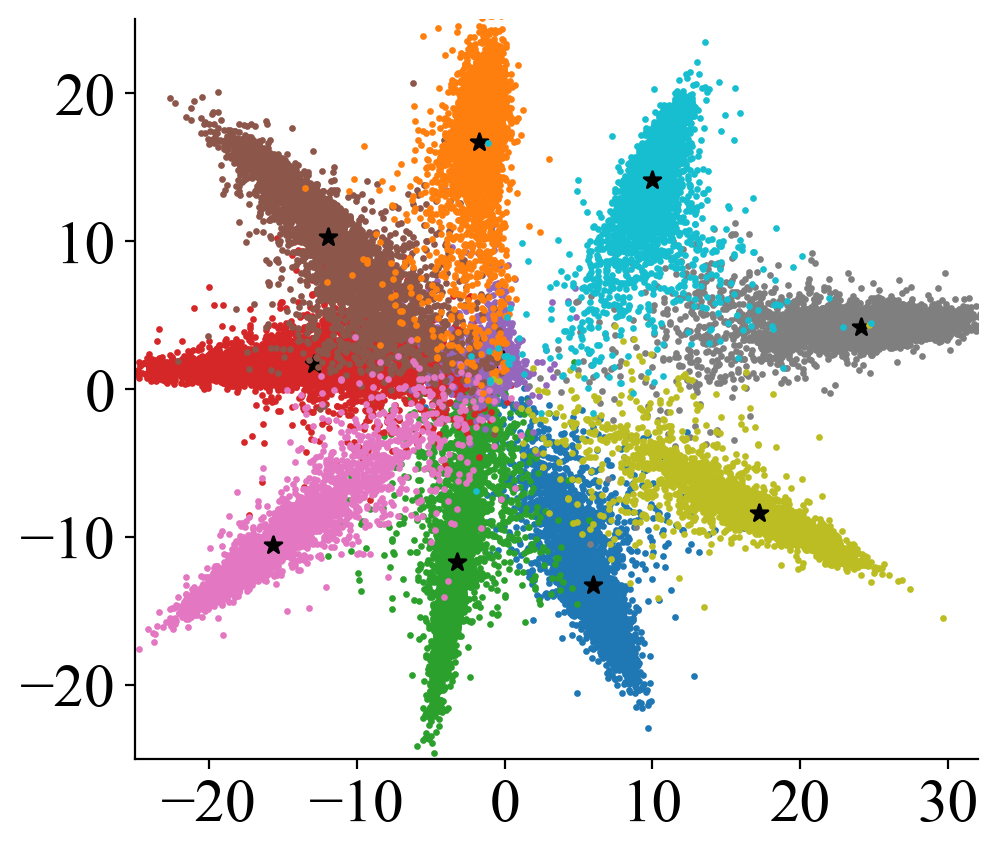}}

\vspace{2.5mm}
\caption{\small \textbf{Embedding features visualization on CIFAR-10.} Teacher and student are ResNet-56 and ResNet-20, respectively. The same color belongs to the same category. $\bigstar$ mean that category center.}
\label{fig:suppl-res56-20} 
\end{figure}
\subsection{More Visualization of Embedding Features.}

Although PCA~\cite{pca} or t-SNE~\cite{tsne} have proven to be effective nonlinear dimensionality reduction techniques, we still adhere to the common practice of providing a more intuitive understanding. Therefore, we follow the approach of~\cite{wen2016discriminative, xu2019larger} and introduce a 2-dimensional learnable feature output at the feature layer for visual analysis. We select the feature statistics of 10 classes from the teacher and student models on CIFAR-10 and visualize their 2-dimensional features, as shown in Fig.~\ref{fig:suppl-res56-20}. Our approach, KD++, clearly demonstrates more intuitive results.

\subsection{Multiple Experiments With Error Bars.}

\begin{figure}[h]
\centering
    \resizebox{1.0\textwidth}{!}{%
    \subcaptionbox{KD++}
    {\includegraphics[width=0.32\textwidth]{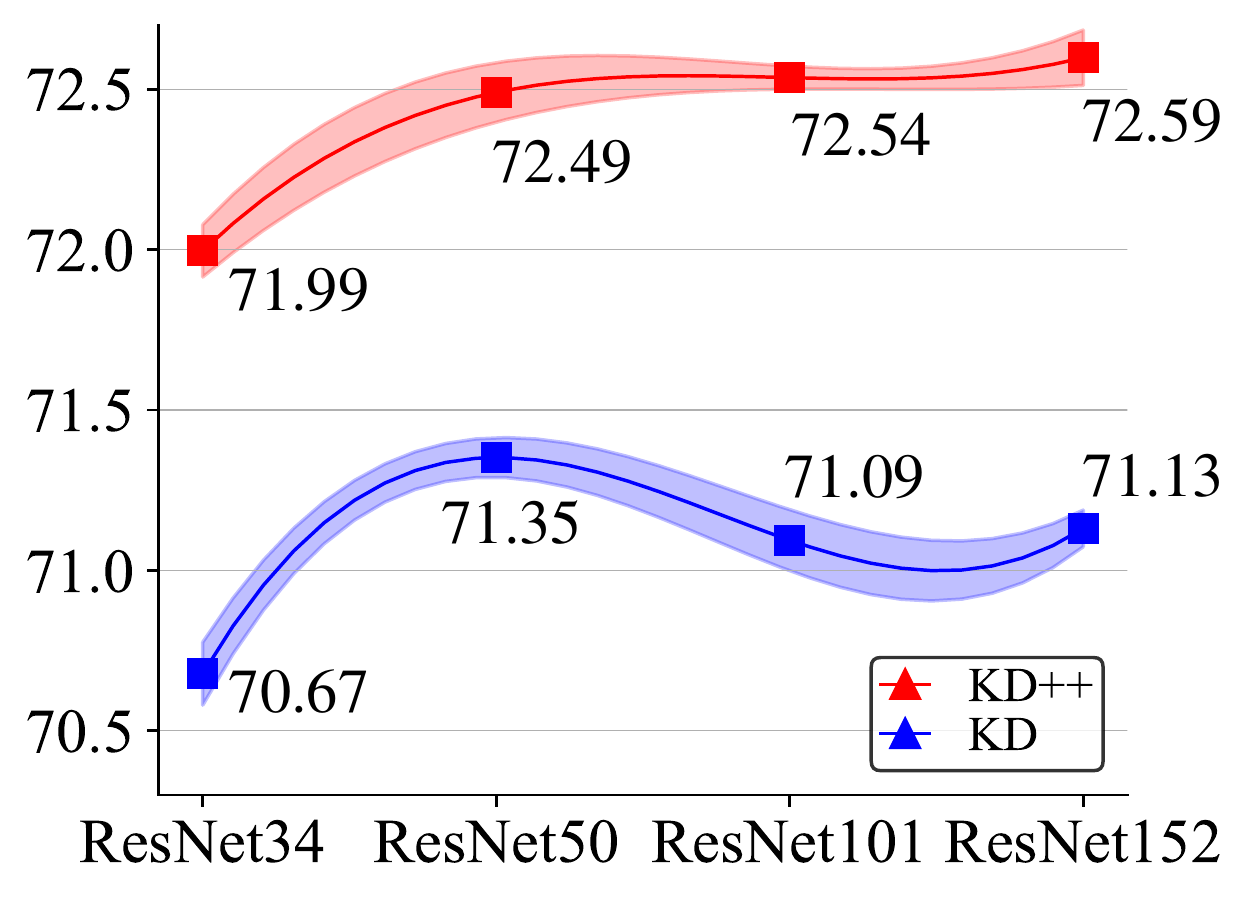}}
    \hfill
    \subcaptionbox{DKD++}
    {\includegraphics[width=0.32\textwidth]{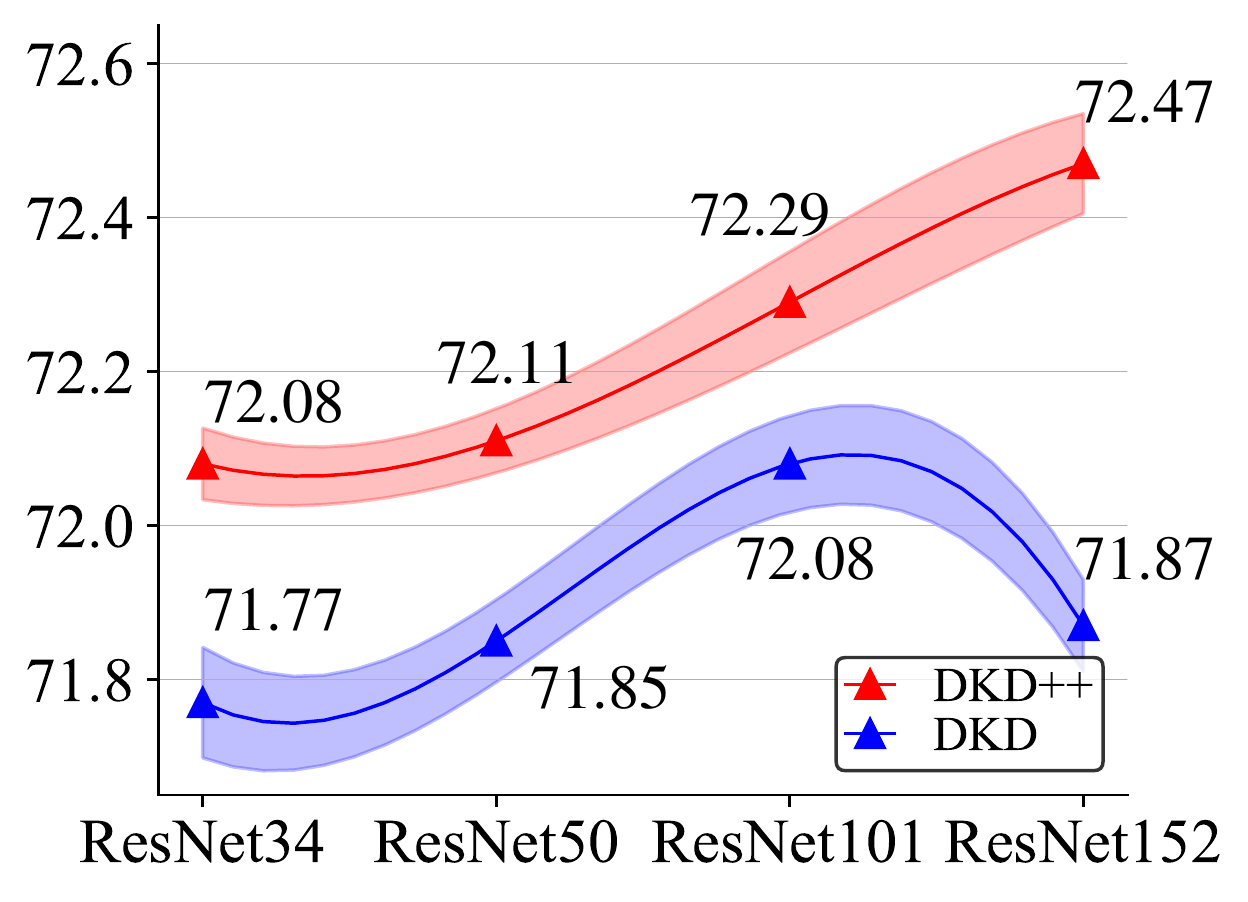}}
    \hfill
    \subcaptionbox{ReviewKD++}
    {\includegraphics[width=0.32\textwidth]{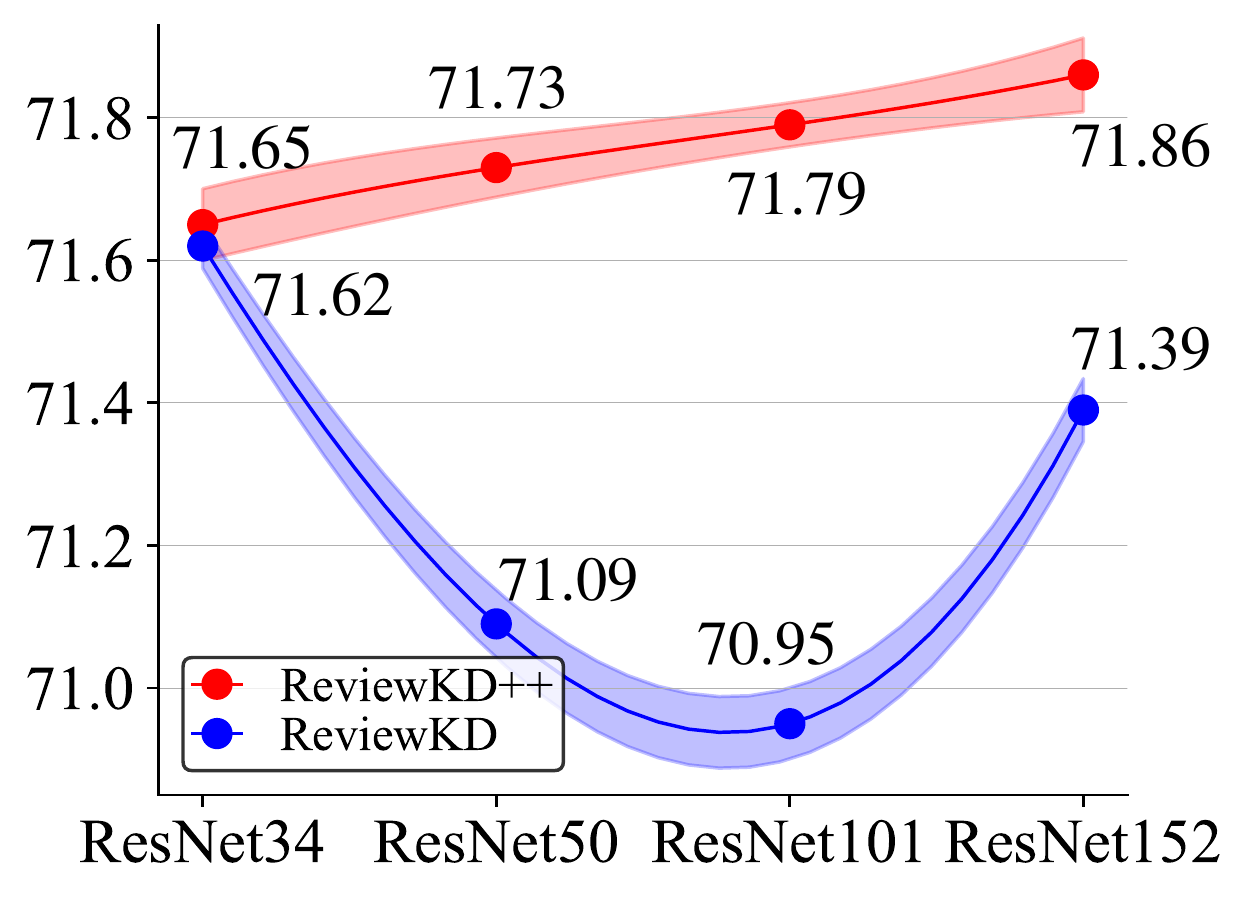}}
    }

\caption{\small
\textbf{Our method can benefit from larger teachers.} With the teacher capacity increasing, our method, KD++, DKD++ and ReviewKD++ (\textcolor{red}{\textbf{red}}) is able to learn better distillation results, even though the original distillation frameworks (\textcolor{blue}{\textbf{blue}}) suffers from degradation problems. The student is ResNet-18, when scaling up the teacher from ResNet-34 to ResNet-152, and reported the Top-1 accuracy ($\%$) on the ImageNet validation set. All results are the average over 5 trials.
}
\vspace{-2.0mm}
\label{fig:suppl_err_bar}
\end{figure}

We accomplish multiple experiments on KD++, DKD++, and ReviewKD++ with increasing teacher model sizes to assess the stability. The results, shown in Fig.~\ref{fig:suppl_err_bar} with error bars, clearly demonstrate our methodology remains stable throughout multiple trials.
\subsection{The Sample Selection Strategy for Class Mean.}

For small-scale datasets such as CIFAR, we compute the mean of the embedded features of samples in the entire training set as the class centers. In practice, these models often suffer from overfitting, achieving close to 100\% accuracy on the training set. Therefore, using all samples does not affect the class centers. However, for large-scale datasets like ImageNet, the models exhibit lower accuracy on the training set (73\%, for example). In such cases, using all training samples to evaluate class centers would inevitably impact the distribution of each class center. We investigate two methods for computing class centers on ImageNet: (1) utilizing all samples and (2) only considering the correctly predicted samples by the teacher model. It is important to note that all samples are derived from the training set. The teacher and student models are ResNet-34 and ResNet-18, with a teacher accuracy of approximately 73\% on the ImageNet training set. We found that the result (72.01\%) by only the correctly predicted samples by the teacher model slightly outperforms using all samples (71.98\%). This confirms the existence of this issue in large-scale datasets; however, the impact is insignificant. Therefore, we default to using all samples for computing class centers.

\end{document}